\documentclass[letterpaper]{article} 
\usepackage{aaai25}  
\usepackage{times}  
\usepackage{helvet}  
\usepackage{courier}  
\usepackage[hyphens]{url}  
\usepackage{graphicx} 
\usepackage{natbib}  
\usepackage{caption} 
\usepackage{bm}
\usepackage{amssymb}
\usepackage{amsmath}
\usepackage{multirow}
\usepackage{caption}
\usepackage{subcaption}
\usepackage{pifont}
\usepackage{makecell}
\usepackage{booktabs}
\usepackage{hyperref}
\hypersetup{hypertex=true,
colorlinks=true,
linkcolor=blue,
anchorcolor=blue,
citecolor=blue}
\usepackage[table ]{ xcolor}
\newenvironment{proof}{{\noindent\it\textbf{Proof.}}\ }{\hfill $\square$\par}
\usepackage{algorithm}
\usepackage{algorithmic}

\usepackage{newfloat}
\usepackage{listings}
\DeclareCaptionStyle{ruled}{labelfont=normalfont,labelsep=colon,strut=off} 
\floatstyle{ruled}
\newfloat{listing}{tb}{lst}{}
\floatname{listing}{Listing}
\newcommand*\samethanks[1][\value{footnote}]{\footnotemark[#1]}

\title{Long-Tailed Out-of-Distribution Detection: Prioritizing Attention to Tail}
\author{
    Yina He\textsuperscript{\rm 1},
    Lei Peng\textsuperscript{\rm 1},
    Yongcun Zhang\textsuperscript{\rm 1},
    Juanjuan Weng\textsuperscript{\rm 2}\thanks{Corresponding author},
    Shaozi Li\textsuperscript{\rm 1,3},
    Zhiming Luo\textsuperscript{\rm 1,3}\samethanks[1]
}
\affiliations{
    \textsuperscript{\rm 1}Department of Artificial Intelligence, Xiamen University, Xiamen, China\\
    \textsuperscript{\rm 2}College of Information Science and Technology, Jinan University, Guangzhou, China \\
    \textsuperscript{\rm 3}{\small Key Laboratory of Multimedia Trusted Perception and Efficient Computing, Ministry of Education of China, Xiamen University}
}

\begin{document}

\maketitle

\begin{abstract}
Current out-of-distribution (OOD) detection methods typically assume balanced in-distribution (ID) data, while most real-world data follow a long-tailed distribution. Previous approaches to long-tailed OOD detection often involve balancing the ID data by reducing the semantics of head classes. However, this reduction can severely affect the classification accuracy of ID data. The main challenge of this task lies in the severe lack of features for tail classes, leading to confusion with OOD data. To tackle this issue, we introduce a novel Prioritizing Attention to Tail (PATT) method using augmentation instead of reduction. Our main intuition involves using a mixture of von Mises-Fisher (vMF) distributions to model the ID data and a temperature scaling module to boost the confidence of ID data. This enables us to generate infinite contrastive pairs, implicitly enhancing the semantics of ID classes while promoting differentiation between ID and OOD data. To further strengthen the detection of OOD data without compromising the classification performance of ID data, we propose feature calibration during the inference phase. By extracting an attention weight from the training set that prioritizes the tail classes and reduces the confidence in OOD data, we improve the OOD detection capability. Extensive experiments verified that our method outperforms the current state-of-the-art methods on various benchmarks. Code is available at \url{https://github.com/InaR-design/PATT}.

\end{abstract}
\section{Introduction}

\begin{figure}[t!]
  \centering
  \subfloat[Separate ACC for head and tail on ImageNet-LT.]
  {\includegraphics[width=0.95\columnwidth]{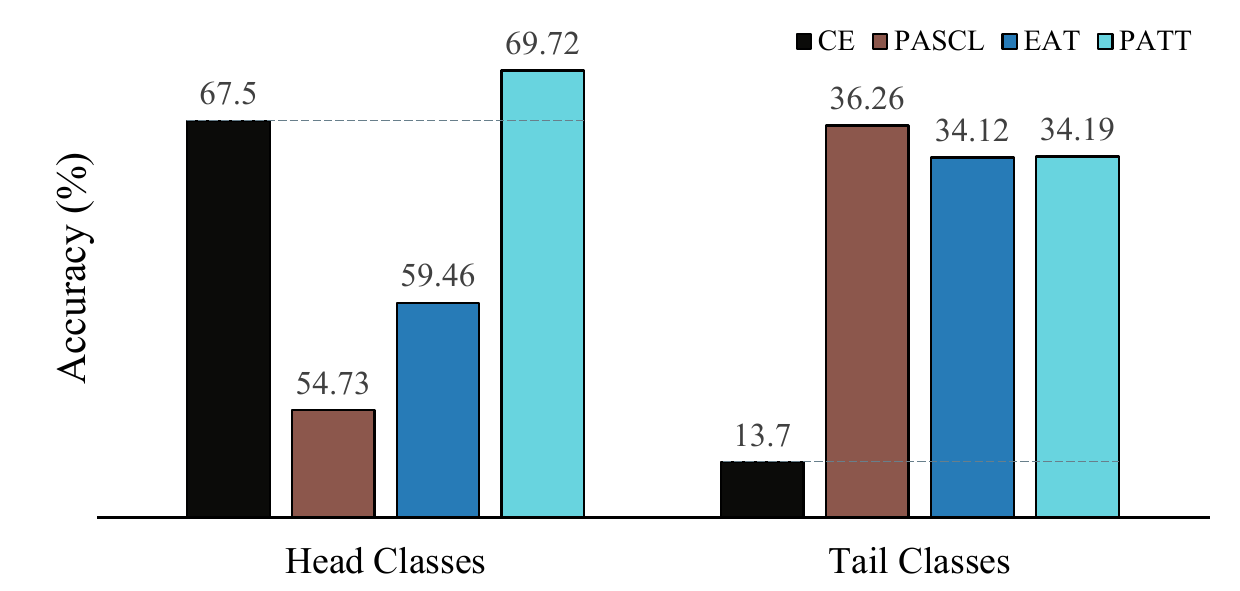}} \\
  \subfloat[PASCL's feature distribution]
  {\includegraphics[width=0.48\columnwidth]{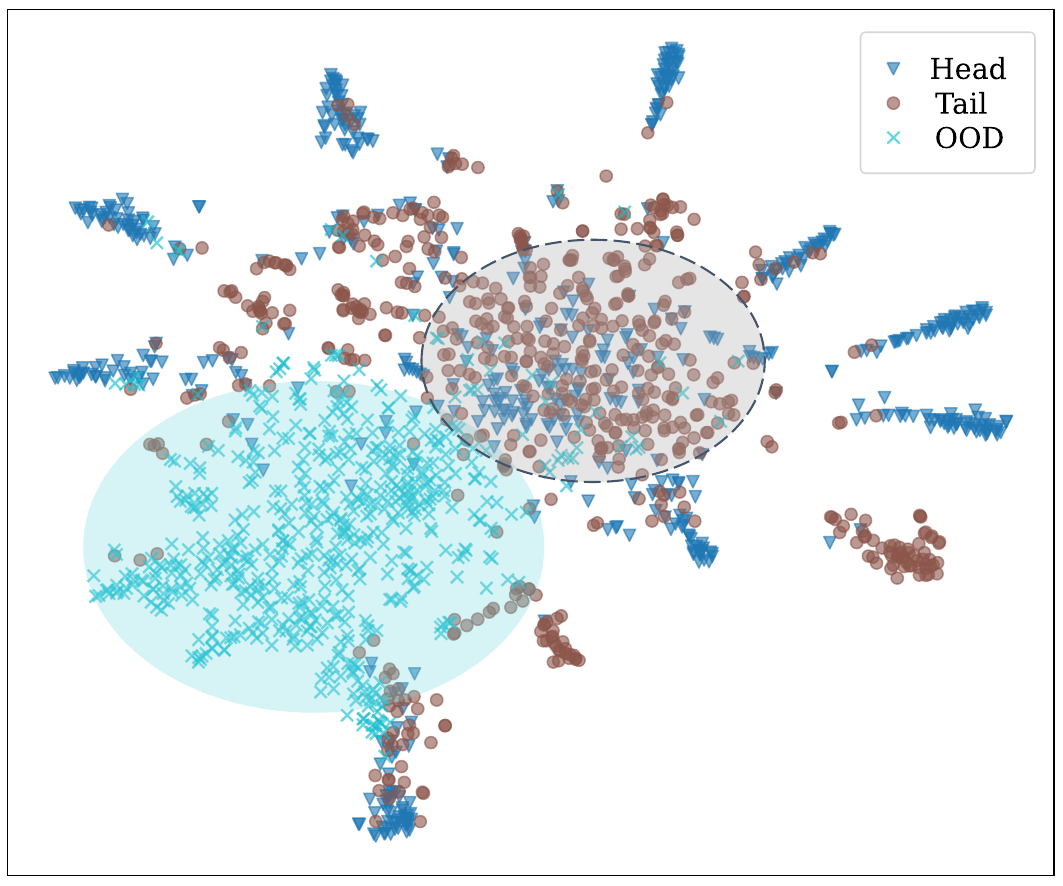}}
  \hfill
  \subfloat[PATT's feature distribution]
  {\includegraphics[width=0.48\columnwidth]{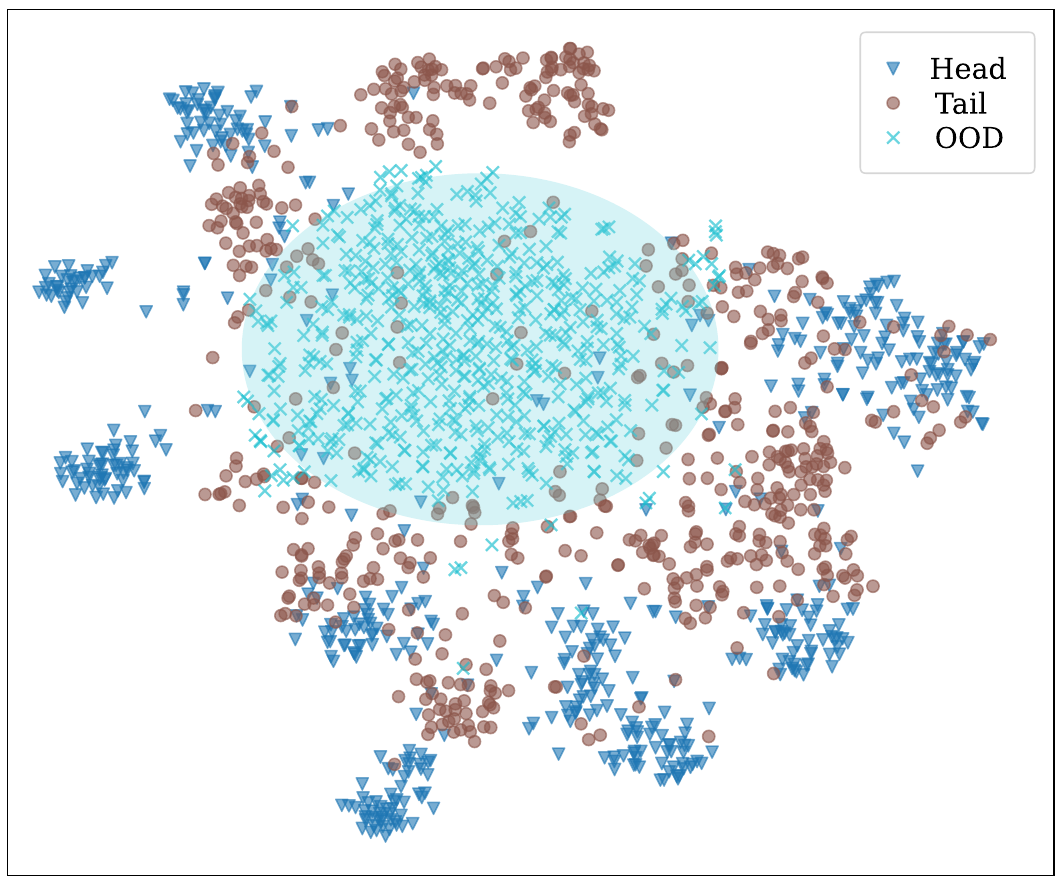}}
  \caption{Visualization of the Comparison between PATT and other methods on ImageNet-LT. (a) Comparison of separate accuracy for head and tail classes on ImageNet-LT between PATT and other methods. (b) (c) Visualization of feature distribution across the top ten classes (Head), the bottom ten classes (Tail), and OOD data from PASCL and PATT.}
  \label{motivate}
\end{figure}


When confronted with a sample that does not match any of the known classes, deep neural networks (DNNs) tend to predict this OOD sample as one of the training classes with high confidence~\cite{hendrycks2017baseline, hein2019relu, hsu2020generalized}. To address this issue, 
numerous OOD detection methods have been proposed and achieved significant improvements~\cite{hendrycks2018deep, liu2020energy, mohseni2020self}. Current state-of-the-art methods introduce surrogate OOD datasets during training, enabling the model to see beyond the training set~\cite{zhudiversified, mingexploit}.
These methods tackle the OOD detection task by maximizing uncertainty~\cite{hendrycks2018deep} and using informative extrapolation based on surrogate outliers~\cite{zhudiversified}.
However, most of these approaches assume a balanced ID data, a condition not typically hold in real-world scenarios characterized by long-tail distributions (e.g., Cybersecurity~\cite{yi2021improved} and Autonomous Driving~\cite{kendall2017uncertainties}). 

In long-tailed recognition, the data distribution is highly imbalanced, and combining long-tailed recognition with OOD detection methods still fails to achieve optimal results~\cite{wang2022partial}. Recent methods like PASCL~\cite{wang2022partial} and EAT~\cite{wei2024eat} focus on differentiating tail classes from OOD data by handling head and tail classes differently. However, these methods significantly reduce ID classification accuracy compared to long-tailed learning approaches. To identify the cause of the reduction, Fig.~\ref{motivate} (a) compares the accuracy of head and tail classes on ImageNet-LT~\cite{liu2019large} across different methods, where CE stands for cross-entropy. We can observe that after incorporating the OOD detection task, both PASCL and EAT show a decline in head class accuracy.
For further analysis, Fig.~\ref{motivate} (b) illustrates PASCL's feature distribution, where head classes appear slightly overfitted, with tail classes, head classes, and OOD data all mixed together, providing insights into the cause of the decline in head class accuracy.

Motivated by these observations, we propose a Prioritizing Attention to Tail (PATT) method for long-tailed OOD detection. The core idea of our design is: address the imbalance in ID data and subsequently enhance the distinction in confidence levels between ID and OOD samples. Specifically, we propose a temperature scaling-based implicit semantic augmentation contrastive learning (TISAC). Compared to traditional supervised contrastive learning~\cite{khosla2020supervised}, TISAC incorporates two key insights tailored for long-tailed OOD detection: (1) \textit{implicit semantic augmentation contrastive learning} effectively balances imbalanced ID data. Particularly, it models the ID data using a mixture of von Mises-Fisher (vMF) distributions, allowing us to sample a large number of contrastive pairs. Yet, sampling sufficient data from the vMF distributions at each training iteration is still inefficient, and we extend the number of samples to infinity and rigorously derive a closed-form for the expected contrastive loss. (2) \textit{temperature scaling}, which ensures high confidence in ID data, thereby increasing the confidence gap between ID and OOD data. By doing so, our method improves accuracy for both head and tail classes as shown in Fig.~\ref{motivate} (a), where CE is a pure classification model, and the other three methods are designed for long-tailed OOD tasks. 

To further boost the long-tailed OOD detection, we propose Post-Hoc Feature Calibration, which derives an attention weight from the training set to refine features during the inference phase, yielding a more balanced ID feature distribution and clearly distinguishing ID and OOD data based on confidence. Ultimately, the feature distribution of our method shows that both head and tail classes occupy nearly equal space, and OOD data is similarly distanced from each class as illustrated in Fig.~\ref{motivate} (c). Our key contributions are:

\begin{itemize}
    \item We identified the potential issues in current long-tailed OOD detection methods, requiring a trade-off where improved OOD detection performance comes at the expense of ID classification accuracy.
    \item We propose a novel Prioritizing Attention to Tail (PATT) framework, composed of temperature scaling-based implicitly augmented contrastive learning and post-hoc feature calibration, functioning during the training and inference stages respectively. The former ensures a balanced feature extractor and classifier, while the latter calibrates features to focus more on tail classes.
    \item Extensive experiments were conducted to validate the effectiveness of PATT in improving long-tailed recognition and OOD detection performance, resulting in a 4.29\% increase in AUROC and an 8.35\% increase in ID classification accuracy on ImageNet-LT.
\end{itemize}

\section{Related Work}
\paragraph{OOD Detection}
The OOD detection~\cite{nguyen2015deep} aims to determine whether an input sample belongs to ID classes or OOD classes. Some works enhance performance by generating virtual OOD data, such as DivOE~\cite{zhu2024diversified}, VOS~\cite{du2022vos} and NPOS~\cite{tao2023non}.
These methods adaptively sample virtual outliers from low-likelihood regions to extrapolate the feature distribution of the OOD data and obtain a more reasonable decision boundary. Other works leverage original OOD data, such as Outlier Exposure (OE)~\cite{hendrycks2018deep} and EnergyOE~\cite{liu2020energy}. They exploits information from OOD data by enforcing a uniform distribution to its logits or maximizing the free energy of OOD samples. Additionally, there are post-hoc strategies that focus on designing new OOD scoring functions, which are always used in conjunction with the aforementioned methods, such as MSP, EnergyOE, and ODIN. Despite the high performance of existing OOD detectors, they are typically trained on class-balanced ID datasets and cannot be directly applied to long-tailed tasks.

\paragraph{Long-Tailed Recognition (LTR)}
Early LTR methods primarily involve re-sampling~\cite{buda2018systematic, byrd2019effect, he2009learning, wallace2011class} and re-weighting~\cite{huang2016learning,cui2019class}, which are effective but limited in addressing intra-class diversity in tail classes, leading to overfitting. To increase the intra-class diversity of tail data, many data augmentation methods have been employed, but explicit data augmentation methods~\cite{bowles2018gan, zhang2018mixup, yun2019cutmix} require significant time and resources. ISDA~\cite{wang2019implicit}, an implicit data augmentation method that utilizes a mixture of Gaussian distribution to generate infinite samples, is an excellent solution to the aforementioned issue. Subsequently, RISDA~\cite{chen2022imagine} and ProCo~\cite{du2024probabilistic}emerged, which effectively applies ISDA to long-tailed learning. Although these LTR methods exhibit excellent performance in the long-tailed classification, they lack specific designs for OOD samples.

\paragraph{Long-tailed OOD detection}
Long-tailed OOD detection has garnered increasing attention, and several methods~\cite{jiang2023detecting, wang2022partial, wei2022open, choi2023balanced} have been proposed for this challenging task. Among them, PASCL~\cite{wang2022partial} and BERL~\cite{choi2023balanced} are OE-based methods. Specifically, PASCL optimizes the contrastive objective between tail class samples and OOD data, pushing them apart in the latent feature space. BERL~\cite{choi2023balanced} is a balanced version of EnergyOE~\cite{liu2020energy}. Recent methods~\cite{wei2024eat,miao2024out} utilize Outlier Class Learning, which directly learns outlier classes for OOD data. EAT~\cite{wei2024eat} builds on this by learning multiple outlier classes and employing Cutmix~\cite{yun2019cutmix} augmentation on the tail with OOD data, which encourages the model to focus on the foreground. COCL~\cite{miao2024out} uses a partial contrastive learning approach similar to PASCL and performs logit calibration during inference. However, these methods reduce semantic information for head classes to improve OOD detection performance, while sacrificing the ID classification accuracy.


\begin{figure*}[t]
    \centering
    \includegraphics[width=\linewidth]{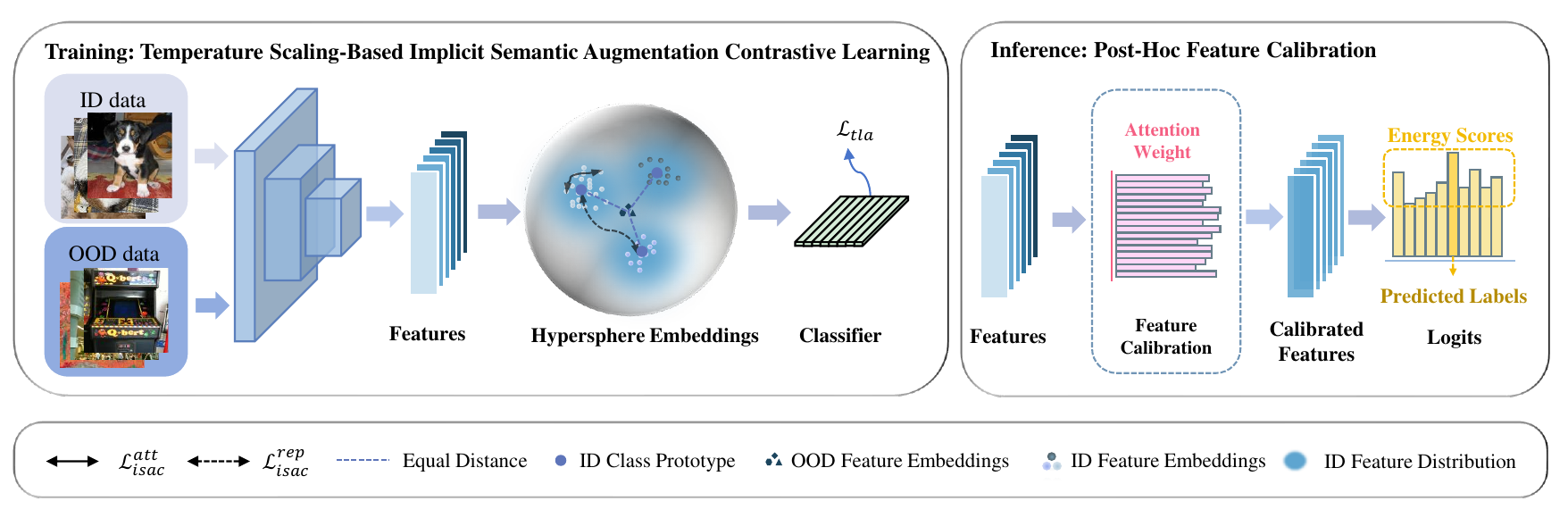}
    \caption{Overview of the proposed framework. The framework consists of a temperature scaling-based implicit semantic augmentation training phase and a feature calibration inference phase. We jointly optimize two complementary terms to encourage desirable hypersphere embeddings: an implicit semantic augmentation contrastive loss to encourage a balanced feature encoder and a temperature scaling-based logit adjustment loss to encourage a balanced high-confidence classifier. Feature calibration fine-tunes features during the inference phase by using an attention weight extracted from the training set, thereby achieving desirable ID classification and OOD detection results.}
    \label{fig:framework}
\end{figure*}

\section{Preliminaries}
\paragraph{Task Definition:}
Following~\cite{wei2024eat, wang2022partial}, let $\mathcal{D}_{in}$ and $\mathcal{D}_{out}$ denote the training set, characterized by a long-tailed ID set, and a surrogate OOD set, respectively. The entire task can be seen as a combination of a multi-class classification for ID data and a binary classification for OOD detection. Let $\mathcal{X} = \mathcal{D}_{in} \cup \mathcal{D}_{out}$ and $Y^{in} = \left\{ 1,2,\dots,K \right\}$ denote the input and label space of the ID data. OOD detection in LTR aims to learn a encoder $f\colon \mathcal{X}\rightarrow \mathcal{Z}$ and a classifier $\varphi\colon \mathcal{Z}\rightarrow \mathcal{Y}$ such that for any test data $x \in \mathcal{X}$: 
if $x$ is drawn from $\mathcal{D}_{in}$, then model can classify $x$ into the correct ID class, otherwise if $x$ is drawn from $\mathcal{D}_{out}$, then model can detect $x$ as OOD data.

\paragraph{Outlier Exposure (OE)}
OE~\cite{hendrycks2018deep} formulates the training objective for OOD detection as follows:
\begin{equation}
    \mathcal {L}_{OE} =  \mathbb {E}_{x,y \sim \mathcal {D}_{in}}[\ell(f(x),y] + \gamma \mathbb {E}_{x \sim \mathcal {D}_{out}}[\ell(f(x),u)], 
    \label{eq:OE}
\end{equation}
where $\gamma$ is a hyperparameter, $u$ represents a uniform distribution, and $ \ell $ denotes the cross entropy loss. We define $\mathcal{L}_{out}=\mathbb {E}_{x \sim \mathcal {D}_{out}}[\ell(f(x),u]$. 
However, CE loss is not an optimal solution for long-tailed recognition, as it treats each sample equally and tends to optimize for those classes that appear more frequently \cite{liu2019large}.

\paragraph{Supervised Contrastive Learning (SCL)}
SCL~\cite{khosla2020supervised} is a form of contrastive learning in a supervised setting, achieving classification by repelling instances from different classes and attracting those from the same class, and has recently emerged as a paradigm for long-tailed recognition. For a feature $z_i$ of class $y$ in a batch $B$, SCL minimize the following loss function:
\begin{equation}
    {\mathcal{L}}^{in}_{scl}(\bm{z}_i,y) = - \log\Bigg\{{\frac{1}{|B_y|}\sum_{p\in B_y}\frac{e^{\bm{z}_i\cdot\bm{z}_p/\tau}}{\sum\limits_{j = 1}^K \sum\limits_{a\in B_j}e^{\bm{z}_i\cdot\bm{z}_a/\tau}}}\Bigg\}
    \label{eq:SCL},
\end{equation}
where $\bm{z}_i$ denotes the feature of $x_i$ drawn from $f$, $B_y$ is a subset of B that contains all instances with the same label $y$, and $|B_y|$ is its cardinality. $\tau>0$ is a scalar temperature hyperparameter that controls tolerance to similar samples; a smaller temperature leads to less tolerance for similar samples~\cite{wang2021understanding}.

\paragraph{Logit Adjustment (LA)} 
LA~\cite{menonlong} aims to refine the model's output probabilities to better match the true probabilities observed in the data. In LTR, it is often used in conjunction with contrastive learning~\cite{zhu2022balanced,hong2021disentangling,tan2020equalization}, the former focuses on representation learning, while the latter learns a balanced classifier. Its definition is as follows:
\begin{equation}
    \label{eq:LA}
    \mathcal{L}_{\text{la}}(\bm{z}_i,y)=-\log\frac{ \pi_{y}e^{ \varphi_{y}(\bm{z}_i)  }}{\sum\limits_{y' \in \mathcal{Y}} \pi_{y'}e^{\varphi_{y'}(\bm{z}_i)}},
\end{equation}
Here, $\pi_{y}=\frac{N_y}{N}$ denotes the class prior of label $y$, $\varphi_{y}(\bm{z}_i)$ is the logits of class $y$ drawn from $\bm{z}_i$.

\section{The Proposed Method}
\label{sec:method}

\subsection{Framework}
The overview of our unified end-to-end model PATT is shown in Fig.~\ref{fig:framework}. It consists of two main components, i.e., \textit{temperature scaling-based implicit semantic augmentation contrastive learning (TISAC)} and \textit{post-hoc feature calibration (FC)}. TISAC is designed to achieve balanced representation learning and classifier for ID data, ensuring high confidence in ID data for better OOD detection capability. FC prioritizes attention to the features of tail classes and reduces the confidence of OOD data, thus benefiting both ID classification and OOD detection. Below, we introduce each component in detail.


\subsection{Temperature Scaling-Based Implicit Semantic Augmentation Contrastive Learning}
Samples from tail classes struggle to represent the diversity of tail data. Additionally, as aforementioned, contrastive learning achieves classification by repelling instances from different classes and attracting same-class ones. Due to the severe imbalance in the training set, the repelling force from the head classes can cause the tail classes to be compressed together, making them difficult to separate. Naive re-weighting and re-sampling methods are difficult to achieve ideal results, as these approaches fail to address the semantic deficiency of tail classes. Inspired by implicit semantic data augmentation methods~\cite{wang2019implicit,chen2022imagine,du2024probabilistic}, we propose a temperature scaling-based implicit semantic augmentation contrastive learning module (TISAC), which consists of two components: Implicit Semantic Augmentation Contrastive Learning (ISAC) and Temperature Scaling-Based Logit Adjustment (TLA). ISAC is inspired by ISDA, which employs Gaussian distributions to model unconstrained features and obtains an upper bound on the expected loss. However, Our method diverges significantly from ISDA-based methods. Compared to Gaussian distribution, the vMF distribution focuses on feature direction rather than magnitude, which largely reduces model bias between classes. Additionally, the Gaussian distribution requires extensive data to compute covariance matrices, making it impractical to estimate Gaussian parameters for all classes with long-tailed data. In contrast, vMF's mean direction vector and concentration parameter can be effectively calculated across batches. Moreover, we rigorously derived the expected loss instead of an upper bound. TLA is a temperature scaling-based version of logit adjustment, which, when combined with OE, creates a larger separation between ID and OOD data.
\subsubsection{Hypersphere Distribution}
Given the desirable properties of hypersphere embeddings, we choose a mixture of von Mises-Fisher (vMF) distributions~\cite{mardia2009directional} to model the ID data. The probability density function of the vMF distribution for a unit vector $\bm{z}\in\mathbb{R}^{d}$ is defined as: 
\begin{equation}
    \label{eq:vMF}
    P_d(\bm{z} ; \bm{\mu_y}, \kappa_y)=Z_{d}(\kappa_y) e^{\kappa_y \bm{\mu_y}^\top\*\bm{z}},\\
\end{equation}
where $\bm{\mu_y}$ is the class prototype with unit norm of class $y$, $\kappa_y\ge 0$ indicates the concentration of the distribution around the mean direction $\bm{\mu_y}$, and $Z_{d}(\kappa_y)$ serving as the normalization factor. A larger $\kappa_y$ means the distribution is more tightly concentrated around the mean direction. Vice versa when $\kappa_y=0$, its turned into a uniform distribution. Under this probability model, we use a mixture of vMF distributions to model the feature distribution: 
\begin{equation}
    \label{eq:vmf}
    P_d(\bm{z}) = \sum_{{y}=1}^{K} P_d(y) P_d(\bm{z}|y) = \sum_{{y}=1}^{K} \pi_{y} Z_{d}(\kappa) e^{\kappa \bm{\mu_y}^\top\*\bm{z}}  ,
\end{equation}
\subsubsection{Implicit Semantic Augmentation Contrastive Learning}
Given the above distribution, an intuitive idea is to obtain contrastive learning pairs by infinitely sampling from the distribution. However, we realize that sampling a sufficient amount of data from the vMF distributions at each training iteration is still inefficient. Therefore, we mathematically derive a closed-form expression when considering an infinite set of training sample pairs, akin to the methods proposed in previous works~\cite{wang2019implicit,du2024probabilistic}. Thus, we have the following formula: 
\begin{equation}
    {\mathcal{L}}_{\rm {isac}}(\bm{z_i},y) = \log \Bigg\{ {\sum\limits_{j = 1}^K \frac{\pi_{j}Z_d(\tilde\kappa_{y})Z_d(\kappa_j)}{\pi_{y}Z_d( \kappa_{y})Z_d(\tilde\kappa_j)} } \Bigg\} , \label{eq:isac}
\end{equation}
where $\tilde \kappa_j = ||\kappa_{j} \bm{\mu}_j +  \bm{z}_i/\tau  ||_2$, which $\bm{z}_j \sim {\rm vMF}(\bm{\mu}_{j}, {\kappa}_{j})$, $\tau$ is a temperature parameter. See proof in Appendix B.

\subsubsection{Temperature Scaling-Based Logit Adjustment}
Through Eq.~\ref{eq:isac}, we can achieve balanced representation learning for ID data. 
Besides, in our task, we further need high confidence in ID data to ensure excellent OOD detection capability. Therefore, we introduce Logit Adjustment~\cite{ding2021local, joy2023sample, kull2019beyond} with a temperature scaling hyperparameter $\varepsilon$ to achieve this goal, which is defined as follows:
\begin{equation}
    \label{eq:TLA}
    \mathcal{L}_{\text{tla}}(\bm{z}_i,y)=-\log\frac{ \pi_{y}e^{ \varphi_{y}(\bm{z}_i)/\varepsilon   }}{\sum\limits_{y' \in \mathcal{Y}} \pi_{y'}e^{\varphi_{y'}(\bm{z}_i)/\varepsilon }}.
\end{equation}

Finally, the overall loss function is as follows:
\begin{equation}
    \mathcal{L}=\mathcal{L}_{\text{isac}} + \alpha \mathcal{L}_{\text{tla}} + \beta \mathcal{L}_{\text{out}},
    \label{eq:total}
\end{equation}
where $\alpha$ and $\beta$ are hyperparameters for each component.

\subsection{Post-Hoc Feature Calibration}
\label{sec:FC}
\subsubsection{Attention Weight}
\begin{figure}[t]
    \centering
    \includegraphics[width=1.0\linewidth]{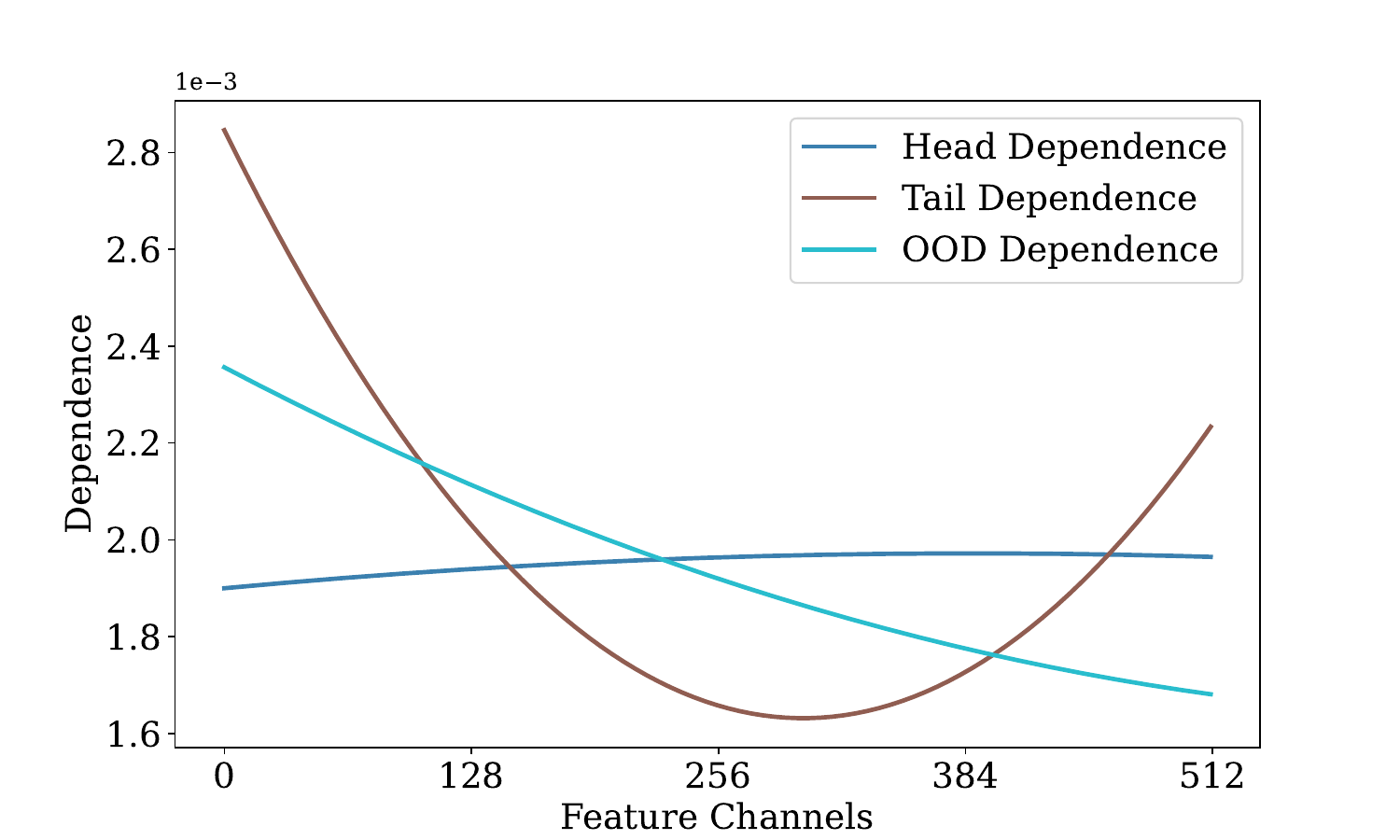}
    \caption{It visualizes the dependence of OOD, head class, and tail class samples on feature channels, showing that these three types of samples rely on different feature channels. }
    \label{fig:fc_motivation}
\end{figure}

The penultimate feature layer is more correlated with the final classification, and different feature channels within it have different correlations with different classes. We further propose to use an attention weight to balance the influence of different feature channels.
Unlike~\cite{tang2023unsupervised}, which measures the importance of feature channels by introducing a learnable weight matrix, we aim to determine which channels are crucial for tail class classification and OOD detection by an attention weight extracted from a class-balanced ID dataset and an OOD dataset, denoted as $\mathcal{D}^{cb}_{in}\subset \mathcal{D}_{in}$ and $\mathcal{D}_{out}$. Let $\mathcal{X}^{cb} = \mathcal{D}^{cb}_{in} \cup \mathcal{D}_{out}$. Given a sample $x_i\in\mathcal{X}^{cb}$, if it comes from $D^{cb}_{in}$, its label is naturally $y_i$. If it comes from $\mathcal{D}_{out}$, its label is determined by the prediction obtained after passing through the network. For convenience, we also refer to this predicted label as $y_i$. Then, we obtain the corresponding $d$-dimensional feature embedding $z_i=f(x_i,\theta)=[z^1_i,z^2_i,\dots,z^d_i]^T$. Then the score of $z_i$ being predicted as class $y_i$ is  $S_{y_i}(z_i)=\varphi(z_i)$. The importance of $k$-th dimension of $z_i$ is defined as $I^k(z_i)=\frac{\partial S_{y_i}(z_i)}{\partial z^k_i} \cdot z^k_i$. We can find that $I(z_i)$ represents the contribution of all channels of $z_i$ to the correct classification. As shown in Fig.~\ref{fig:fc_motivation}, we can obtain that OOD data, head class, and tail class depend on different feature channels. To achieve balanced results for ID data and satisfactory OOD detection performance, it is essential to derive feature-level representations that favor tail classes and enforce a uniform distribution for OOD data. Specifically, we calculate an attention weight $A$ for each feature in the test set for calibration, which is defined as follows:
\begin{equation}
    \label{eq:attention}
    A= \frac{1}{|N|}\sum\limits_{j = 1}^K\bigg\{\sum\limits_{i\in N^{in}_{j}}\frac{I(z^{in}_i)}{\pi_j}-\sum\limits_{i\in N^{out}_{j}}\frac{I(z^{out}_i)}{\pi_j}\bigg\},
\end{equation}
where $z^{in}_i\in \mathcal{D}^{cb}_{in}$, $z^{out}_i\in \mathcal{D}_{out}$ and its label $j$ is a virtual label comes from the prediction obtained after passing through the network, $|N|$ is the cardinality of the class-balanced set $\mathcal{X}^{cb}$. 
Besides, the resulting attention weight $A$ has a large variance, then we scale it to the range [0,2] as $A^{scale}$, where values between [0,1] are attenuated and values between [1,2] are enhanced.

\subsubsection{OOD Score}
 During the inference phase, we element-wise multiply this weight $A^{scale}$ with feature $z$ to obtain the final calibrated feature $z^{cal}=z\odot A^{scale}$. By doing so, our model will prioritize attention to the features of tail classes and reduce the confidence of OOD data. Thus, feature calibration benefits both ID classification and OOD detection. Additionally, instead of using maximum softmax probability (MSP) as OOD scores, we are inspired by energyOE~\cite{liu2020energy} and use energy scores as OOD scores in the inference phase, which is defined as follows:
\begin{equation}
    \label{eq:score}
    {\mathcal{S}}_{\rm {ood}}(\bm{z_i})=\log\sum\limits_{j = 1}^K{e^{\varphi_{j}(\bm{z}^{cal}_i)}}.
\end{equation}

\section{Experiments}
\subsection{Experiment Settings}
\label{sec:settings}
\subsubsection{Datasets:}
We conduct experiments on widely used datasets, i.e., CIFAR10-LT, CIFAR100-LT~\cite{cao2019learning}, and ImageNet-LT~\cite{liu2019large} as ID training sets ($\mathcal{D}_{in}$). The standard CIFAR10, CIFAR100, and ImageNet test sets are used as ID test sets ($D^{test}_{in}$). Following PASCL~\cite{wang2022partial}, we utilize 300,000 samples from TinyImages80M~\cite{torralba200880} as the surrogate OOD training data for CIFAR10/100-LT and ImageNet-Extra as the surrogate OOD training for ImageNet-LT. 
We set the default imbalance ratio to 100 for CIFAR10/100-LT. For OOD test data, we use Textures~\cite{cimpoi2014describing}, SVHN ~\cite{netzer2011reading}, Tiny ImageNet~\cite{le2015tiny}, LSUN~\cite{yu2015lsun}, and Places365~\cite{zhou2017places} introduced in the SC-OOD benchmark~\cite{yang2021semantically} as $\mathcal{D}_{test}^{out}$ for CIFAR10/100-LT. Additionally, for near-OOD experiments~\cite{fort2021exploring}, we use CIFAR-100 as $\mathcal{D}_{test}^{out}$ for CIFAR10-LT and vice versa. We use ImageNet-1k-OOD as $\mathcal{D}_{test}^{out}$ for ImageNet-LT. More information about the datasets can be found in Appendix C.

\begin{table}[t!]
\centering
\begin{subtable}{1\linewidth}
\resizebox{\linewidth}{!}{
\begin{tabular}{cccccc}
\hline
$\mathcal{D}_{\text{out}}^{\text{test}}$ & Method & AUROC$\uparrow$ & AUPR-in$\uparrow$ & AUPR-out$\uparrow$ & FPR95$\downarrow$ \\ 
\hline
\multirow{3}{*}{Texture} & OE & 92.59$_{\pm \text{0.4}}$ & 96.01$_{\pm \text{1.4}}$ & 83.32$_{\pm \text{1.7}}$ & 25.10$_{\pm \text{1.1}}$ \\ 
 & PASCL & 93.16$_{\pm \text{0.4}}$ & 96.57$_{\pm \text{1.2}}$ & 84.80$_{\pm \text{1.5}}$ & \textbf{23.26}$_{\pm \text{0.9}}$ \\ 
 & \textbf{Ours} & \textbf{93.96}$_{\pm \text{0.6}}$ & \textbf{97.69}$_{\pm \text{0.8}}$ & \textbf{86.49}$_{\pm \text{0.8}}$ & 26.65$_{\pm \text{1.6}}$ \\
\hline
\multirow{3}{*}{SVHN} & OE & 95.10$_{\pm \text{1.0}}$ & 91.59$_{\pm \text{0.5}}$ & 97.14$_{\pm \text{0.8}}$ & 16.15$_{\pm \text{1.5}}$ \\
 & PASCL & 96.63$_{\pm \text{0.9}}$ & 92.89$_{\pm \text{0.5}}$ & 98.06$_{\pm \text{0.6}}$ & 12.18$_{\pm \text{3.3}}$ \\ 
 & \textbf{Ours} & \textbf{98.21}$_{\pm \text{0.7}}$ & \textbf{97.19}$_{\pm \text{0.6}}$ & \textbf{98.50}$_{\pm \text{0.5}}$ & \textbf{5.73}$_{\pm \text{2.0}}$\\
\hline
\multirow{3}{*}{\small{CIFAR100}} & OE & 83.40$_{\pm \text{0.3}}$ & 84.06$_{\pm \text{0.3}}$ & 80.93$_{\pm \text{0.6}}$ & 56.96$_{\pm \text{0.9}}$ \\
 & PASCL & 84.43$_{\pm \text{0.2}}$ & 85.32$_{\pm \text{0.5}}$ & 82.99$_{\pm \text{0.5}}$ & 57.27$_{\pm \text{0.9}}$ \\ 
 & \textbf{Ours} & \textbf{85.36}$_{\pm \text{0.2}}$ & \textbf{86.01}$_{\pm \text{0.3}}$ & \textbf{83.29}$_{\pm \text{0.5}}$ & \textbf{51.12}$_{\pm \text{0.9}}$ \\
\hline
\multirow{3}{*}{\shortstack{Tiny \\ ImageNet}} & OE & 86.14$_{\pm \text{0.3}}$ & 89.88$_{\pm \text{0.7}}$ & 79.33$_{\pm \text{0.7}}$ & 47.78$_{\pm \text{0.7}}$ \\
 & PASCL & 87.14$_{\pm \text{0.2}}$ & 90.22$_{\pm \text{0.5}}$ & 81.54$_{\pm \text{0.4}}$ & 47.69$_{\pm \text{0.6}}$ \\ 
 & \textbf{Ours} & \textbf{88.62}$_{\pm \text{0.2}}$ & \textbf{90.82}$_{\pm \text{0.7}}$ & \textbf{84.54}$_{\pm \text{0.1}}$ & \textbf{41.30}$_{\pm \text{1.7}}$ \\
\hline
\multirow{3}{*}{LSUN} & OE & 91.35$_{\pm \text{0.2}}$ & 93.06$_{\pm \text{0.3}}$ & 87.62$_{\pm \text{0.8}}$ & 27.86$_{\pm \text{0.7}}$ \\
 & PASCL & \textbf{93.17}$_{\pm \text{0.15}}$ & 82.59$_{\pm \text{0.3}}$ & 91.76$_{\pm \text{0.5}}$ & 26.40$_{\pm \text{1.0}}$ \\ 
 & \textbf{Ours} & 91.64$_{\pm \text{0.3}}$ & \textbf{93.16}$_{\pm \text{0.6}}$ & \textbf{92.27}$_{\pm \text{0.1}}$ & \textbf{24.41}$_{\pm \text{1.5}}$ \\
\midrule
\multirow{3}{*}{Place365} & OE & 90.07$_{\pm \text{0.3}}$ & 82.09$_{\pm \text{0.4}}$ & 95.15$_{\pm \text{0.2}}$ & 34.04$_{\pm \text{0.9}}$ \\
 & PASCL & 91.43$_{\pm \text{0.2}}$ & 82.59$_{\pm \text{0.2}}$ & 96.28$_{\pm \text{0.1}}$ & 33.40$_{\pm \text{0.9}}$ \\ 
 & \textbf{Ours} & \textbf{91.95}$_{\pm \text{0.6}}$ & \textbf{82.63}$_{\pm \text{0.3}}$ & \textbf{97.81}$_{\pm \text{0.2}}$ & \textbf{30.15}$_{\pm \text{1.9}}$ \\
\hline
\multirow{3}{*}{Average} & OE & 89.77$_{\pm \text{0.3}}$ & 89.45$_{\pm \text{0.4}}$ & 87.25$_{\pm \text{0.6}}$ & 34.65$_{\pm \text{0.5}}$ \\
 & PASCL & 90.99$_{\pm \text{0.2}}$ & 90.18$_{\pm \text{0.4}}$ & 89.24$_{\pm \text{0.3}}$ & 33.36$_{\pm \text{0.8}}$ \\ 
 & \textbf{Ours} & \textbf{91.62}$_{\pm \text{0.5}}$ & \textbf{91.25}$_{\pm \text{0.5}}$ & \textbf{90.48}$_{\pm \text{0.3}}$ & \textbf{29.89}$_{\pm \text{1.4}}$ \\
\hline
\end{tabular}
}
\caption{\centering{Comparison of PATT to PASCL and OE on six OOD datasets.
}}
\label{cifar10_table_top}

\resizebox{\linewidth}{!}{
\begin{tabular}{cccccc}
\hline
Method & AUROC & AUPR-in & AUPR-out & FPR95 & ACC \\ 
\hline
MSP & 72.28 & 73.96 & 70.27 & 66.07 & 72.34\\
EnergyOE & 89.31 & 91.01 & 88.92 & 40.88 & 74.68\\
OE & 89.77$_{\pm \text{0.3}}$ & 89.45$_{\pm \text{0.4}}$ & 87.25$_{\pm \text{0.6}}$ & 34.65$_{\pm \text{0.5}}$ & 73.84$_{\pm \text{0.8}}$\\
PASCL & 90.99$_{\pm \text{0.2}}$ & 90.18$_{\pm \text{0.4}}$ & 89.24$_{\pm \text{0.3}}$ & 33.36$_{\pm \text{0.8}}$ & 77.08$_{\pm \text{1.0}}$\\
EAT & \textbf{91.73}$_{\pm \text{0.3}}$ & \textbf{91.46}$_{\pm \text{0.5}}$ & \underline{90.35}$_{\pm \text{0.5}}$ & \underline{30.30}$_{\pm \text{1.2}}$ & \underline{81.31}$_{\pm \text{0.3}}$\\
\hline
\textbf{Ours} & \underline{91.62}$_{\pm \text{0.5}}$ & \underline{91.25}$_{\pm \text{0.5}}$ & \textbf{90.48}$_{\pm \text{0.3}}$ & \textbf{29.89}$_{\pm \text{1.4}}$ & \textbf{84.77}$_{\pm \text{0.2}}$ \\
\hline
\end{tabular}
}
\caption{\centering{Comparison results with different competing methods. The results are averaged over the six OOD test datasets in (a). }}
\label{cifar10_table_down}
\end{subtable}
\caption{Comparison results on CIFAR10-LT. The best results are shown in bold, and the second-best results are underlined.}
\label{table_cifar10}
\end{table}

\subsubsection{Evaluation Metrics:}
Following ~\citet{wang2022partial} and  \citet{yang2021semantically}, we use four metrics to evaluate OOD detection and ID classification: (1) \textbf{AUROC} ($\uparrow$) is the area under the receiver operating characteristic curve from OOD scores; (2) \textbf{AUPR} ($\uparrow$) is the area under precision-recall curve. AUPR contains AUPR-in which ID samples are treated as positive and AUPR-out is vice versa; (3) \textbf{FPR95} ($\downarrow$) is the false positive rate (FPR) when 95\% OOD samples have been successfully detected; (4) \textbf{ACC} ($\uparrow$) is the classification accuracy of the ID data.

\subsubsection{Configuration:}
Following PASCL~\cite{wang2022partial}, we use ResNet-18~\cite{he2016deep} as our backbone and perform the experiments using the Adam optimizer~\cite{kingma2014adam} with an initial learning rate $1 \times 10^{-3}$ for experiments on CIFAR10/100-LT. For ImageNet-LT, we use ResNet-50~\cite{he2016deep} as our backbone and train the model using the SGD optimizer with an initial learning rate of 0.1. All experiments are conducted by training the model for 100 epochs, with a batch size of 128. The reported results for OE, PASCL, and EAT are presented as the mean and standard deviation over six runs. More detailed configuration information is presented in Appendix C, and the full algorithm of our PATT is described in Appendix A.

\begin{table}[t!]
\centering
\begin{subtable}{1\linewidth}
\centering
\resizebox{\linewidth}{!}{
\begin{tabular}{cccccc}
\hline
$\mathcal{D}_{\text{out}}^{\text{test}}$ & Method & AUROC$\uparrow$ & AUPR-in$\uparrow$ & AUPR-out$\uparrow$ & FPR95$\downarrow$ \\ 
\hline
\multirow{3}{*}{Texture} & OE & 76.71$_{\pm \text{1.2}}$ & 85.28$_{\pm \text{1.0}}$ & 58.79$_{\pm \text{1.4}}$ & 68.28$_{\pm \text{1.5}}$ \\ 
 & PASCL & 76.01$_{\pm \text{0.7}}$ & 85.84$_{\pm \text{1.0}}$ & 58.12$_{\pm \text{1.1}}$ & 67.43$_{\pm \text{1.9}}$ \\ 
 & \textbf{Ours} & \textbf{76.86}$_{\pm \text{0.7}}$ & \textbf{85.86}$_{\pm \text{0.9}}$ & \textbf{59.16}$_{\pm \text{0.9}}$ & \textbf{66.64}$_{\pm \text{1.4}}$ \\
\hline
\multirow{3}{*}{SVHN} & OE & 77.61$_{\pm \text{3.3}}$ & 73.25$_{\pm \text{1.5}}$ & 86.82$_{\pm \text{2.5}}$ & 58.04$_{\pm \text{4.8}}$ \\
 & PASCL & 80.19$_{\pm \text{2.2}}$ & 67.81$_{\pm \text{2.4}}$ & 88.49$_{\pm \text{1.6}}$ & 53.45$_{\pm \text{3.6}}$ \\ 
 & \textbf{Ours} & \textbf{90.21}$_{\pm \text{2.5}}$ & \textbf{84.66}$_{\pm \text{1.3}}$ & \textbf{92.49}$_{\pm \text{1.7}}$ & \textbf{32.12}$_{\pm \text{3.2}}$ \\
\hline
\multirow{3}{*}{\shortstack{CIFAR- \\ 10}} & OE & 62.23$_{\pm \text{0.3}}$ & 66.16$_{\pm \text{0.4}}$ & 57.57$_{\pm \text{0.3}}$ & 80.64$_{\pm \text{1.0}}$ \\
 & PASCL & 62.33$_{\pm \text{0.4}}$ & 67.21$_{\pm \text{0.2}}$ & 57.14$_{\pm \text{0.2}}$ & 79.55$_{\pm \text{0.8}}$ \\ 
 & \textbf{Ours} & \textbf{63.12}$_{\pm \text{0.5}}$ & \textbf{67.69}$_{\pm \text{0.3}}$ & \textbf{60.77}$_{\pm \text{0.3}}$ & \textbf{78.89}$_{\pm \text{0.4}}$ \\
\hline
\multirow{3}{*}{\shortstack{Tiny \\ ImageNet}} & OE & 68.04$_{\pm \text{0.4}}$ & 79.36$_{\pm \text{0.3}}$ & 51.66$_{\pm \text{0.5}}$ & 76.66$_{\pm \text{0.5}}$ \\
 & PASCL & 68.20$_{\pm \text{0.4}}$ & 79.65$_{\pm \text{0.4}}$ & 51.53$_{\pm \text{0.4}}$ & 76.11$_{\pm \text{0.8}}$ \\ 
 & \textbf{Ours} & \textbf{71.02}$_{\pm \text{0.2}}$ & \textbf{80.94}$_{\pm \text{0.6}}$ & \textbf{56.57}$_{\pm \text{0.7}}$ & \textbf{75.42}$_{\pm \text{1.4}}$ \\
\hline
\multirow{3}{*}{LSUN} & OE & 77.10$_{\pm \text{0.6}}$ & 85.33$_{\pm \text{0.6}}$ & 61.42$_{\pm \text{1.0}}$ & 63.98$_{\pm \text{1.4}}$ \\
 & PASCL & 77.19$_{\pm \text{0.4}}$ & 85.73$_{\pm \text{0.5}}$ & 61.27$_{\pm \text{0.7}}$ & 63.31$_{\pm \text{0.9}}$ \\ 
 & \textbf{Ours} & \textbf{78.46}$_{\pm \text{0.9}}$ & \textbf{86.24}$_{\pm \text{0.5}}$ & \textbf{65.79}$_{\pm \text{0.9}}$ & \textbf{59.86}$_{\pm \text{1.2}}$ \\
\midrule
\multirow{3}{*}{Place365} & OE & 75.80$_{\pm \text{0.5}}$ & 60.99$_{\pm \text{0.6}}$ & 86.68$_{\pm \text{0.4}}$ & 65.72$_{\pm \text{0.9}}$ \\
 & PASCL & 76.02$_{\pm \text{0.2}}$ & 60.84$_{\pm \text{0.4}}$ & 86.52$_{\pm \text{0.3}}$ & 64.81$_{\pm \text{0.3}}$ \\ 
 & \textbf{Ours} & \textbf{77.85}$_{\pm \text{0.6}}$ & \textbf{61.65}$_{\pm \text{0.5}}$ & \textbf{87.45}$_{\pm \text{0.9}}$ & \textbf{64.70}$_{\pm \text{1.5}}$ \\
\hline
\multirow{3}{*}{Average} & OE & 72.91$_{\pm \text{0.7}}$ & 75.06$_{\pm \text{0.6}}$ & 67.16$_{\pm \text{0.6}}$ & 68.89$_{\pm \text{1.1}}$ \\
 & PASCL & 73.56$_{\pm \text{0.3}}$ & 74.52$_{\pm \text{0.4}}$ & 67.18$_{\pm \text{0.1}}$ & 67.44$_{\pm \text{0.6}}$ \\ 
 & \textbf{Ours} & \textbf{76.25}$_{\pm \text{0.9}}$ & \textbf{77.84}$_{\pm \text{0.5}}$ & \textbf{70.37}$_{\pm \text{0.9}}$ & \textbf{62.94}$_{\pm \text{1.4}}$ \\
\hline
\end{tabular}
}
\caption{\centering{Comparison of PATT to PASCL and OE on six OOD datasets.
}}
\label{cifar100_table_top}

\resizebox{\linewidth}{!}{
\begin{tabular}{cccccc}
\hline
Method & AUROC & AUPR-in & AUPR-out & FPR95 & ACC \\ 
\hline
MSP & 61.00 & 64.52 & 57.54 & 82.01 & 40.97\\
EnergyOE & 71.10 & 75.42 & 67.23 & 71.78 & 39.05\\
OE & 72.91$_{\pm \text{0.7}}$ & 75.06$_{\pm \text{0.6}}$ & 67.16$_{\pm \text{0.6}}$ & 68.89$_{\pm \text{1.1}}$ & 39.04$_{\pm \text{0.4}}$\\
PASCL & 73.32$_{\pm \text{0.3}}$ & 74.52$_{\pm \text{0.4}}$ & 67.18$_{\pm \text{0.1}}$ & 67.44$_{\pm \text{0.6}}$ & 43.10$_{\pm \text{0.5}}$\\
EAT & \underline{74.41}$_{\pm \text{0.9}}$ & \underline{76.04}$_{\pm \text{0.7}}$ & \underline{69.57}$_{\pm \text{0.9}}$ & \underline{65.05}$_{\pm \text{1.5}}$ & \underline{46.13}$_{\pm \text{0.3}}$\\
\hline
\textbf{Ours} & \textbf{76.25}$_{\pm \text{0.9}}$ & \textbf{77.84}$_{\pm \text{0.5}}$ & \textbf{70.37}$_{\pm \text{0.9}}$ & \textbf{62.94}$_{\pm \text{1.4}}$ & \textbf{50.07}$_{\pm \text{0.3}}$ \\
\hline
\end{tabular}
}
\caption{\centering{Comparison results with different competing methods. The results are averaged over the six OOD test datasets in (a).}}
\label{cifar100_table_down}
\end{subtable}
\caption{Comparison results on CIFAR100-LT. The best results are shown in bold, and the second-best results are underlined.}

\label{table_cifar100}
\end{table}

\begin{table}[t!]
\centering
\resizebox{\linewidth}{!}{
\begin{tabular}{cccccc}
\hline
Method & AUROC$\uparrow$ & AUPR-in$\uparrow$ & AUPR-out$\uparrow$ & FPR95$\downarrow$ & ACC$\uparrow$ \\ 
\hline
MSP & 53.81 & 37.68 & 51.63 & 90.15 & 39.65\\
EnergyOE & 64.76 & 44.63 & 64.77 & 87.72 & 38.50 \\
OE & 66.33 & 43.17 & 68.29 & 89.96 & 40.13\\
PASCL & 68.00 & 44.32 & 70.15 & 87.53 & 45.49\\
EAT & 69.84 & 46.67 & 69.25 & 87.63 & 46.79 \\
\hline
\textbf{Ours} & \textbf{74.13} & \textbf{51.41} & \textbf{87.43} & \textbf{80.57} & \textbf{55.14} \\
\hline
\end{tabular}
}
\caption{Comparison results on ImageNet-LT with ImageNet-1k-OOD as OOD test dataset.}
\label{ImageNet_table}
\end{table}

\begin{table*}[t!]
\centering
\begin{tabular}{c|ccc|cccccc}
\hline
ID Dataset & ISAC & TLA & FC & AUROC$\uparrow$ & AUPR-in$\uparrow$ & AUPR-out$\uparrow$ & FPR95$\downarrow$ & ACC$\uparrow$ & ACC-t$\uparrow$ \\
\hline
\hline
\multirow{6}{*}{CIFAR10-LT} 
 & \ding{55} & \ding{55} & \ding{55} & 71.21 & 74.84 & 64.37 & 58.09 & 78.10 & 63.00 \\
 & \ding{51} & \ding{55} & \ding{55} & 74.10 & 78.70 & 66.43 & 49.09 & 79.55 & 66.47 \\
 & \ding{55} & \ding{51} & \ding{55} & 84.70 & 83.68 & 81.12 & 48.18 & 81.11 & 70.77 \\
 & \ding{55} & \ding{55} & \ding{51} & 79.75 & 80.48 & 74.38 & 52.17 & 79.55 & 67.50 \\
 & \ding{51} & \ding{51} & \ding{55} & 91.06 & 90.90 & 88.92 & 32.35 & 82.09 & 70.00 \\
 & \multicolumn{3}{c|}{\cellcolor[gray]{0.85}\textbf{PATT}} & \cellcolor[gray]{0.85}\textbf{91.62} & \cellcolor[gray]{0.85}\textbf{91.25} & \cellcolor[gray]{0.85}\textbf{90.48} & \cellcolor[gray]{0.85}\textbf{29.89} & \cellcolor[gray]{0.85}\textbf{84.77} & \cellcolor[gray]{0.85}\textbf{79.67} \\
\hline
\hline
\multirow{6}{*}{CIFAR100-LT}
 & \ding{55} & \ding{55} & \ding{55} & 71.70 & 72.77 & 65.95 & 72.22 & 44.68 & 10.97 \\
 & \ding{51} & \ding{55} & \ding{55} & 74.43 & 76.80 & 67.50 & 64.91 & 45.49 & 17.00 \\
 & \ding{55} & \ding{51} & \ding{55} & 72.10 & 73.88 & 66.35 & 69.80 & 49.01 & 27.12 \\
 & \ding{55} & \ding{55} & \ding{51} & 72.95 & 74.43 & 66.51 & 70.79 & 44.90 & 12.72 \\
 & \ding{51} & \ding{51} & \ding{55} & 75.48 & 77.19 & 68.87 & 64.25 & 49.56 & 27.12 \\
 & \multicolumn{3}{c|}{\cellcolor[gray]{0.85}\textbf{PATT}} & \cellcolor[gray]{0.85}\textbf{76.25} & \cellcolor[gray]{0.85}\textbf{77.84} & \cellcolor[gray]{0.85}\textbf{70.37} & \cellcolor[gray]{0.85}\textbf{62.94} & \cellcolor[gray]{0.85}\textbf{50.07} & \cellcolor[gray]{0.85}\textbf{31.03} \\
\hline
\hline
\multirow{6}{*}{ImageNet-LT} 
 & \ding{55} & \ding{55} & \ding{55} & 67.87 & 44.28 & 69.74 & 88.76 & 45.13 & 9.24 \\
 & \ding{51} & \ding{55} & \ding{55} & 68.39 & 45.31 & 79.39 & 83.39 & 48.77 & 13.79 \\
 & \ding{55} & \ding{51} & \ding{55} & 72.74 & 51.17 & 86.25 & 82.08 & 50.27 & 31.58 \\
 & \ding{55} & \ding{55} & \ding{51} & 72.07 & 47.84 & 82.07 & 82.36 & 46.17 & 14.01 \\
 & \ding{51} & \ding{51} & \ding{55} & 73.77 & 50.10 & 87.32 & 81.92 & 55.06 & 33.17 \\
 & \multicolumn{3}{c|}{\cellcolor[gray]{0.85}\textbf{PATT}} & \cellcolor[gray]{0.85}\textbf{74.13} & \cellcolor[gray]{0.85}\textbf{51.41} & \cellcolor[gray]{0.85}\textbf{87.43} & \cellcolor[gray]{0.85}\textbf{80.57} & \cellcolor[gray]{0.85}\textbf{55.14} & \cellcolor[gray]{0.85}\textbf{34.19} \\
\hline
\end{tabular}
\caption{Ablation results of three key modules for PATT on CIFAR10-LT, CIFAR100-LT and ImageNet-LT.}
\label{ablation_table}
\end{table*}

\begin{table}[t!]
\centering
\resizebox{\linewidth}{!}{
\begin{tabular}{ccc|ccccc}
\hline
Baseline & Method & AUROC & AUPR-in & AUPR-out & FPR95 & ACC \\ 
\hline
\multirow{4}{*}{OE~~+} &none &72.91 & 75.06 & 67.16 & 68.89 & 39.04\\
&$\tau$-norm & 73.14 & 74.63 & 66.48 & 68.76 & 40.08\\
&LA & 73.05 & 74.36 & 66.35 & 69.24 & 39.16 \\
&\textbf{FC} & \textbf{74.86} & \textbf{76.34} & \textbf{68.50} & \textbf{67.28} & \textbf{40.24} \\
\hline
\multirow{4}{*}{TISAC~~+} &none & 75.48 & 77.19 & 68.87 & 64.25 & 49.56\\
&$\tau$-norm & 74.39 & 76.29 & 67.86 & 65.08 & 47.83\\
&LA & 73.96 & 75.63 & 67.41 & 66.41 & 41.18 \\
&\textbf{FC} & \textbf{76.25} & \textbf{77.84} & \textbf{70.37} & \textbf{62.94} & \textbf{50.07} \\
\hline
\end{tabular}
}
\caption{Ablation study of feature calibration on CIFAR100-LT using ResNet18.}
\label{table:FC}
\end{table}


\subsection{Comparison with Other Methods}
We compare our method with several leading long-tailed OOD detection methods, including classical methods MSP~\cite{hendrycks2017baseline},  OE~\cite{hendrycks2018deep}, and its variants EnergyOE~\cite{liu2020energy}, PASCL~\cite{wang2022partial} and EAT~\cite{wei2024eat}. Following EAT, we mainly compare the experimental results with OE and PASCL.
For a fair comparison, some results in this paper are directly taken from~\cite{wang2022partial}. For AUPR-in, which is not reported in PASCL, and for some incomplete results, we strictly reproduced them under the same setting. Results on CIFAR10-LT, CIFAR100-LT and ImageNet-LT are presented in Table~\ref{table_cifar10}, Table~\ref{table_cifar100} and Table~\ref{ImageNet_table}.

Table~\ref{cifar10_table_top} and Table~\ref{cifar100_table_top} present a comparison between our method with OE and PASCL on CIFAR10-LT and CIFAR100-LT using six OOD datasets as we mentioned before. Our method consistently outperforms OE and PASCL on six OOD datasets, except for FPR95 of Texture on CIFAR10-LT. Both our method and PASCL use OE as a baseline and employ contrastive learning as the primary approach. Our method implicitly enhances the semantic information of tail classes, while PASCL's partial and asymmetric reduce the semantic information of head classes. Thus, our method reduces the average FPR95 by 3.47\% on CIFAR10-LT and 4.50\% on CIFAR100-LT compared to PASCL, demonstrating a significant improvement.

Table~\ref{cifar10_table_down}, Table~\ref{cifar100_table_down} and Table~\ref{ImageNet_table} report the comparison of PATT to state-of-the-art Long-tailed OOD methods. We can observe that as the dataset size increases, our method can distinguish itself more from other methods. For instance, in CIFAR10-LT, the OOD detection results of PATT and EAT are similar, and PATT even falls short of EAT in terms of AUROC and AUPR-in. However, in CIFAR100-LT and ImageNet-LT, our method greatly outperforms EAT across all metrics. On CIFAR100-LT, our method improves average AUROC by 1.84\% and ID classification accuracy by 3.94\% over EAT. The improvement is even more remarkable on ImageNet-LT, with an increase of 4.29\% in AUROC and 8.35\% in ID classification accuracy. This indicates that our method possesses a more comprehensive semantic representation, making it suitable for real-world scenarios.
\subsection{Ablation Study}
\subsubsection{Analysis of key modules in PATT}
As described in the Method section, our PATT consists of Implicit Semantic Augmentation Contrastive Learning (ISAC), Temperature Scaling-Based Logit Adjustment (TLA), and Feature Calibration (FC). Table~\ref{ablation_table} shows the effectiveness of these three modules on all three ID datasets, reporting the average performance across six OOD datasets. Rows without the ISAC module use SCL to better highlight the importance of ISAC. From this table, we can infer that (1) All three modules enhance the model's OOD performance and ID classification accuracy across all three datasets; (2) TLA achieves a balanced classifier while ensuring high-confidence classification results, significantly improves the AUROC for OOD detection and the ACC of tail classes; (3) The combination of ISAC and TLA maximizes the effectiveness of both modules, as ISAC ensures a balanced feature extractor and TLA achieves a balanced classifier; (4) Adding FC not only enhances OOD detection capability but also improves classification accuracy; (5) While FC provides substantial benefits when used alone, its gains are less pronounced when the model is already relatively balanced.

\subsubsection{On post-hoc feature calibration}
Post-hoc feature calibration calibrates all features using an attention weight extracted from the training set. In this section, we compare post-hoc feature calibration with post-hoc logit adjustment~\cite{menonlong} and $\tau$-norm~\cite{kangdecoupling}. The results are shown in Table~\ref{table:FC}. As can be seen, post-hoc feature calibration, whether combined with the imbalanced feature encoder OE or the balanced encoder TISAC, consistently enhances the OOD detection and ID classification. In contrast, LA and $\tau$-norm only provide slight improvements with the imbalanced encoder and can even lead to worse results when combined with a balanced encoder. A reasonable explanation is that the attention weight obtained from the training set comprehensively summarizes the entire dataset, which cannot be achieved simply by using class priors for debiasing or blindly regularizing the classifier.
\begin{table}[t]
\centering
\begin{tabular}{ ccc }
\hline
\multirow{2}{*}{\textbf{Method}} & \multicolumn{2}{c}{\textbf{ACC ($\uparrow$)}} \\
&  Head classes & Tail classes \\
\hline
OE & 54.29 & 20.90 \\
PASCL & 54.73 (+0.44) & 36.26 (+15.36) \\
EAT & 59.46 (+5.17) & 34.12 (+13.22) \\
Ours & 69.72 (+15.43) & 34.19 (+13.29) \\
\hline
\end{tabular}
\caption{Separate ACC for head and tail on ImageNet-LT. }
\label{tab:classwise}

\end{table}
\subsubsection{Improvements on head and tail classes}
In Table~\ref{tab:classwise}, we show the ACC gains of our method compared to OE on both head and tail classes. It can be seen that our method significantly improves the ACC for both head and tail classes, with an increase of 15.43\% for head classes and 13.29\% for tail classes. Compared to PASCL and EAT, our enhancements are balanced for both head and tail classes. PASCL, in particular, is extremely biased towards tail classes, providing almost no gain for head classes.

\section{Conclusion}
We propose an advanced method for long-tailed OOD detection called Prioritizing Attention to Tail (PATT). PATT implicitly enhances ID data using a mixture of vMF distribution, significantly improving the performance of long-tailed classification. Additionally, we introduce temperature scaling-based logit adjustment, which, combined with Outlier Exposure (OE), creates a large confidence margin between ID and OOD data, enabling the model to identify OOD data effectively. To address biases at important feature levels, we propose an attention-based feature calibration, applied during the inference phase. Compared to post-hoc methods that directly apply debiasing at the logit level, our approach is more comprehensive. Extensive experiments demonstrate that our proposed PATT significantly improves both OOD detection and ID classification performance.
\section*{Acknowledgements}
This work is supported by the National Natural Science Foundation of China (No.~62276221, No.~62376232); the Fujian Provincial Natural Science Foundation of China (No.~2022J01002).

\bibliography{aaai25}

\clearpage

\section*{Appendix}
\section{A. The PATT Algorithm}
\begin{algorithm}[h]
\caption{PATT}
\label{alg:algorithm}
\textbf{Input}: training dataset $ \mathcal {D}_{in}^{train} $; surrogate dataset $ \mathcal {D}_{out}^{train} $; \\ unlabeled dataset $ \mathcal {D}_{in \cup out}^{test} $ \\
\textbf{Training}
\begin{algorithmic}[1] 
\FOR{each iteration} 
\STATE Sample a mini-batch of ID training data, \\
$ \left\{ (x_i^{in}, y_i) \right\}_{i=1}^n $ from $ \mathcal {D}_{in}^{train} $
\STATE Sample a mini-batch of surrogate OOD data, \\
$ \left\{ (x_i^{out}) \right\}_{i=1}^n $ from $ \mathcal {D}_{out}^{train} $
\STATE Perform common gradient descent on feature extractor $f$ and decoder $\varphi$ with $\mathcal{L}$ from Eq. 8
\ENDFOR
\end{algorithmic}
\textbf{Inference}
\begin{algorithmic}[1] 
\FOR{each sample $x$ in dataset $ \mathcal {D}_{in \cup out}^{test} $} 
\STATE Obtain $z$ by inferencing $x$ on feature extractor $f$,
\STATE Obtain $z^{cal}$ by feature calibration $z\odot A^{scale}$,
\STATE Inference $z^{cal}$ by decoder $\varphi$
\ENDFOR
\end{algorithmic}
\end{algorithm}
\section{B. Proof of Implicitly Augmented \\ Supervised Contrastive Learning}
\begin{proof}
    We can obtain the following equation from Eq. 2 by multiplying both the numerator and denominator by the same coefficient: 
    \begin{equation}
        {\mathcal{L}}^{in}_{scl}(\bm{z}_i,y) = \log{|B_y|}-\log{\frac{\sum\limits_{p\in B_y}\frac{|B_y|}{|B|}\frac{1}{|B_y|}e^{\bm{z}_i\cdot\bm{z}_p/\tau}}{\sum\limits_{j = 1}^K \sum\limits_{a\in B_j}\frac{|B_j|}{|B|}\frac{1}{|B_j|}e^{\bm{z}_i\cdot\bm{z}_a/\tau}}}
        \label{eq:scl_v2},
    \end{equation}
    where constant $\log{|B_y|}$ can be omitted in the loss function and $|B_j|$ is the sampling number of class j. Thus $\lim_{|B| \to \infty} |B_j|/|B| = \pi_j$. Then we have the following loss function:
    \begin{equation}
        {\mathcal{L}}^{in}_{scl}(\bm{z}_i,y) = -\log{\frac{\pi_{y}\mathbb{E}[ e^{\bm{z}_i\cdot \tilde{\bm{z}}_{y}/\tau}]}{\sum\limits_{j = 1}^K \pi_{j} \mathbb{E}[ e^{\bm{z}_i\cdot \tilde{\bm{z}}_j/\tau}]}}
        \label{eq:scl_v3},
    \end{equation}
    Now, we introduce the moment-generating function of the von Mises-Fisher (vMF) distribution as follows:
    \begin{equation}
        \mathbb {E} \left(e^{\mathbf {t} ^{\mathrm {T} }\mathbf {z} }\right) = \frac{Z_d(\kappa)}{Z_d(\tilde \kappa)} . 
        \label{eq:moment}
    \end{equation}
    By substituting Eq.~\ref{eq:moment} into Eq.~\ref{eq:scl_v3}, we can obtain:
    \begin{equation}
        {\mathcal{L}}^{in}_{scl}(\bm{z}_i,y) = -\log { \frac{\pi_{y} Z_d( \kappa_{y})}{Z_d(\tilde\kappa_{y})} }  + \log  {\sum\limits_{j = 1}^K  \frac{\pi_j Z_d( \kappa_j)}{Z_d(\tilde\kappa_j)} }. 
        \label{eq:scl_v4},
    \end{equation}
    Finally, we obtain the following loss function:
    \begin{equation}
    {\mathcal{L}}_{\rm {isac}}(\bm{z_i},y) = \log \Bigg\{ {\sum\limits_{j = 1}^K \frac{\pi_{j}Z_d(\tilde\kappa_{y})Z_d(\kappa_j)}{\pi_{y}Z_d( \kappa_{y})Z_d(\tilde\kappa_j)} } \Bigg\} , \label{eq:isac_proof}
    \end{equation}
\end{proof}

\section{C. Configuration Details}
For experiments on CIFAR10/100-LT, we train our model on ResNet18 via the algorithm described in Appendix A for 100 epochs using Adam~\cite{kingma2014adam} optimizer with initial learning rate $1 \times 10^{-3}$. We decay the learning rate to 0 using a cosine annealing learning rate schedule~\cite{loshchilov2016sgdr} with weight decay of $5 \times 10^{-4}$. Our surrogate OOD dataset is a subset of TinyImages80M with 300K images, following PASCL~\cite{wang2022partial} and EAT~\cite{wei2024eat}. TinyImages80M is a large-scale, diverse dataset containing $32 \times 32$ natural images. ~\cite{hendrycks2018deep} selected 300,000 samples from this dataset, ensuring no overlap with the CIFAR datasets. 

For experiments on ImageNet-LT, we train our model on ResNet50 via the algorithm described in Appendix A for 100 epochs using SGD~\cite{loshchilov2016sgdr} optimizer with an initial learning rate of 0.1 and decay the learning rate by a factor of 10 at epoch 60 and 80. Batch size is set to 32 for ID data and 64 for surrogate OOD data. Our surrogate OOD dataset is a subset of ImageNet22k~\cite{deng2009imagenet}, named ImageNet-Extra constructed by PASCL~\cite{wang2022partial}. Specifically, ImageNet-Extra has 517,711 images belonging to 500 classes which is randomly sampled from ImageNet-22k~\cite{deng2009imagenet} but no overlap with ImageNet-LT. Following~\cite{wang2022partial}, we use ImageNet-1k-OOD as test OOD dataset. This dataset has 50,000 images randomly sampled from 1000 classes of ImageNet-22k. To ensure fairness in OOD detection, its size is identical to that of the ID test set. Of course, the 1,000 classes in ImageNet-1k-OOD do not overlap with the 1,000 ID classes in ImageNet-LT and the 500 OOD training classes in ImageNet-Extra.

For experiments on CIFAR10/100-LT and ImageNet-LT, RandAug~\cite{cubuk2020randaugment} is employed as a data augmentation strategy for the classification, and SimAug~\cite{chen2020simple} for contrastive learning. 
Empirically, we set $\varepsilon=0.7$, $\tau=0.1$, $\alpha=0.5$ for both CIFAR10/100-LT and ImageNet-LT, $\beta=0.1$ for CIFAR10/100-LT and  $\beta=0.02$ for ImageNet-LT. We train the above models with Nvidia GeForce 3090Ti GPU

\section{D. Additional Experimental Results}

\subsection{D.1 Additional Ablation Study}

\begin{table}[!h]
\centering
\setlength{\tabcolsep}{1mm}
\begin{tabular}{ ccccc }
\toprule
 {Method} & {AUROC$\uparrow$} & {AUPR-in$\uparrow$} & {AUPR-out$\uparrow$}& {FPR95$\downarrow$} \\
\midrule
\midrule
 ODIN & 73.97 & 51.02 & 87.32 & 80.91  \\
 MSP & 70.72 & 47.90 & 84.50 & 81.07 \\
 \textbf{Energy} & \textbf{74.13} & \textbf{51.41} & \textbf{87.43} & \textbf{80.57}\\
\midrule
\bottomrule
\end{tabular}
\caption{Results on ImageNet-LT with different OOD scoring methods}
\label{tab:ablation_scoring}
\end{table}
\subsubsection{OOD Scoring Methods}
As mentioned in the related work section, OOD scoring methods focus on designing new OOD scoring functions. They are often combined with training-time methods to identify OOD data better. Table~\ref{tab:ablation_scoring} presents an ablation study of three OOD scoring methods: Energy~\cite{liu2020energy}, MSP~\cite{hendrycks2017baseline}, and ODIN~\cite{liang2018enhancing}. The table shows that our method's performance remains relatively stable across these methods and still significantly outperforms EAT when using MSP. This indicates that our performance gains depend not solely on a good OOD scoring method but on the combination with our training-time method.

\subsubsection{Imbalance Ratio $\rho$}
In the Experiments section, we use a default imbalance ratio of $\rho=100$ for both CIFAR10/100-LT and ImageNet-LT. Here, we demonstrate our model's performance under different imbalance ratios by conducting experiments on CIFAR10-LT with $\rho=50$. The results, shown in Table~\ref{tab:cifar100-0.02-ResNet18}, indicate that our method significantly outperforms PASCL~\cite{wang2022partial} and current SOTA method EAT~\cite{wei2024eat} when $\rho=50$.

\begin{table}[!h]
\centering
\setlength{\tabcolsep}{0.7mm}
\begin{tabular}{ cccccc }
\toprule
{Method} & {AUROC} & {AUPR-in} &{AUPR-out}& {FPR95} & {ACC} \\
\midrule
\midrule
 PASCL & 74.70 & 76.10 & 68.45 & 65.98 & 47.99 \\
 EAT & 76.61 & 78.56 & 71.02 & 61.25 &50.11  \\
 Ours & \textbf{77.70}& \textbf{79.60} & \textbf{71.23} & \textbf{59.52} & \textbf{54.49} \\
\midrule
\bottomrule
\end{tabular}
\caption{Average results over six OOD test dataset on CIFAR100-LT ($\rho=50$) using ResNet18.}
\label{tab:cifar100-0.02-ResNet18}
\end{table}

\subsubsection{Model Structure} 
In the Experiments section, we use a standard ResNet18 as the backbone model for CIFAR10/100-LT. Here, we demonstrate our model's performance under more complicated model ResNet34~\cite{he2016deep}. The results, shown in Table~\ref{tab:cifar100-0.01-ResNet34}, indicate that our method outperforms PASCL and EAT by a considerable margin on ResNet34. Furthermore, by comparing Table~2~(b) and Table~\ref{tab:cifar100-0.01-ResNet34}, we find that only our method achieves better results on ResNet34, indicating that the semantics extracted by our method contain more comprehensive information. 

\begin{table}[!h]
\centering

\setlength{\tabcolsep}{0.7mm}
\begin{tabular}{ cccccc }
\toprule
 {Method} & {AUROC} & {AUPR-in} &{AUPR-out}& {FPR95} & {ACC} \\
\midrule
\midrule
 PASCL & 71.49 & 73.67 & 64.45 & 68.72 & 40.41 \\
 EAT & 73.22 & 75.23 & 65.44 & 65.78 & 45.35 \\
 Ours & \textbf{76.80}& \textbf{78.73} & \textbf{70.35} & \textbf{61.49} & \textbf{51.82} \\
\midrule
\bottomrule
\end{tabular}

\caption{Average results over six OOD test dataset on CIFAR100-LT ($\rho=100$) using ResNet34.}
\label{tab:cifar100-0.01-ResNet34}
\end{table}

\subsubsection{Temperature $\tau$}
In this section, we conduct ablation experiments on the temperature $\tau$, which is a hyperparameter in implicitly augmented supervised contrastive learning. The results are shown in Table~\ref{tab:ablation_tau}. We can see that the performance of our method is stable concerning different temperature hyperparameter $\tau$ values, and the best result is obtained when $\tau=0.1$.

\begin{table}[!h]
\centering

\setlength{\tabcolsep}{1mm}
\begin{tabular}{ cccccc }
\toprule
 \textbf{$\tau$} & {AUROC} & {AUPR-in} &{AUPR-out}& {FPR95} & {ACC} \\
\midrule
\midrule
 0.02 & 72.50 & 74.42 & 65.99 & 67.90 & 44.00 \\
 0.05 & 75.64 & 77.71 & 69.50 & 63.06 & 48.27 \\
 0.07 & 75.80 & 77.59 & 69.42 & 62.98 & 49.44 \\
 \textbf{0.1} & \textbf{76.25} & \textbf{77.84} & \textbf{70.37} & \textbf{62.94} & 50.07\\
 0.2 & 74.36 & 76.18 & 68.13 & 66.59 & \textbf{50.14}\\
\midrule
\bottomrule
\end{tabular}
\caption{Average results over six OOD test dataset on CIFAR100-LT with different temperature $\tau$.}
\label{tab:ablation_tau}
\end{table}

\subsubsection{Temperature Scaling $\varepsilon$}
In this section, we conduct ablation experiments on the temperature scaling $\varepsilon$, a hyperparameter in temperature scaling-based logit adjustment. The results are shown in Table~\ref{tab:ablation_ts}. We can see that the performance of our method is stable concerning different temperature scaling hyperparameter $\varepsilon$ values, and the best result is obtained when $\varepsilon=0.7$.

\begin{table}[!h]
\centering

\setlength{\tabcolsep}{1mm}
\begin{tabular}{ cccccc }
\toprule
 {$\varepsilon$} & {AUROC} & {AUPR-in} &{AUPR-out}& {FPR95} & {ACC} \\
\midrule
\midrule
 0.3 & 72.78 & 74.48 & 66.48 & 67.90 & 47.16 \\
 0.5 & 74.70 & 77.00 & 67.76 & 63.93 & \textbf{50.29}\\
 \textbf{0.7} & \textbf{76.25} & \textbf{77.84} & \textbf{70.37} & \textbf{62.94} & 50.07\\
 0.9 & 76.09 & 77.78 & 70.07 & 63.14 & 48.45\\
\midrule
\bottomrule
\end{tabular}
\caption{Average results over six OOD test dataset on CIFAR100-LT with different temperature scaling hyperparameter $\varepsilon$.}
\label{tab:ablation_ts}
\end{table}

\begin{table}[!h]
\centering

\setlength{\tabcolsep}{1mm}
\begin{tabular}{ cccccc }
\toprule
 {IB} & {AUROC} & {AUPR-in} &{AUPR-out}& {FPR95} & {ACC} \\
\midrule
\midrule
 $[0,1]$ & 89.66 & 89.96	& 88.43	& 30.81	& \textbf{84.80} \\
 $[0,2]$ & \textbf{91.62} & \textbf{91.25} & \textbf{90.48} & \textbf{29.89}	& 84.77\\
\midrule
\bottomrule
\end{tabular}
\caption{Average results over six OOD test dataset on CIFAR10-LT with different interval boundary.}
\label{tab:ablation_ib}
\end{table}

\begin{table*}[t!]

\centering
\begin{tabular}{c|ccccc|ccccc}
\toprule
& \multicolumn{5}{c|}{ACC} &\multicolumn{5}{c}{AUROC} \\ 
\midrule
\midrule
\multirow{2}{*}{Hyperparameter $\alpha$} & \multicolumn{5}{c|}{Hyperparameter $\beta$}& \multicolumn{5}{c}{Hyperparameter $\beta$} \\ 

  & 0.01 & 0.02 & 0.05 & 0.1 & 0.2 & 0.01 & 0.02 & 0.05 & 0.1 & 0.2 \\ 
\midrule
0.2   & 47.24 & 47.69 & 47.53 & 49.12 & 48.43 & \textbf{77.28} & 76.53 & 76.72 & 76.34 & 75.98\\ 

0.5   & 48.50 & 47.52 & 48.58 & \textbf{50.07} & 49.60& 76.19 & 75.19 & 75.45 & 76.25 & 75.21\\ 

1   & 49.69 & 48.58 & 49.02 & 48.67 & 48.88 & 74.83 & 75.45 & 75.36 & 75.20 & 74.83 \\ 

\midrule
\bottomrule
\end{tabular}
\caption{Accuracy on CIFAR10-LT using ResNet18 depending on hyperparameter $\gamma$, $\alpha$, and $\beta$. The results are averaged over six OOD test datasets in the SC-OOD benchmark}
\label{hyper_ACC}
\end{table*}

\subsubsection{Hyperparameter $\alpha$ and $\beta$}
As demonstrated in Eq.~8 the hyperparameter $\alpha$ and $\beta$ correspond to the coefficients of the temperature scaling-based logit adjustment and outlier exposure losses, respectively. In this section, we conduct ablation experiments on these parameters, with the results shown in Table~\ref{hyper_ACC}. From Table~\ref{hyper_ACC}, we observe that a smaller $\alpha$ achieves better OOD detection performance, as indicated by the AUROC metric, although the accuracy of ID classification decreases. Overall, the model's performance does not significantly vary with changes in these parameters, indicating that our method achieves excellent results without requiring precise hyperparameter tuning and is highly robust. In terms of parameter optimization, we observe that with $\alpha=0.5$ and $\beta=0.1$, we achieve the best Accuracy, but not the best AUROC. The optimal AUROC is obtained with $\alpha=0.2$ and $\beta=0.01 $, but the Accuracy is the worst one. Therefore, we chose $\alpha=0.5$ and $\beta=0.1$ as parameters for CIFAR10/100-LT.

\begin{figure}[!t]
\centering
\subfloat{
		\includegraphics[scale=0.17]{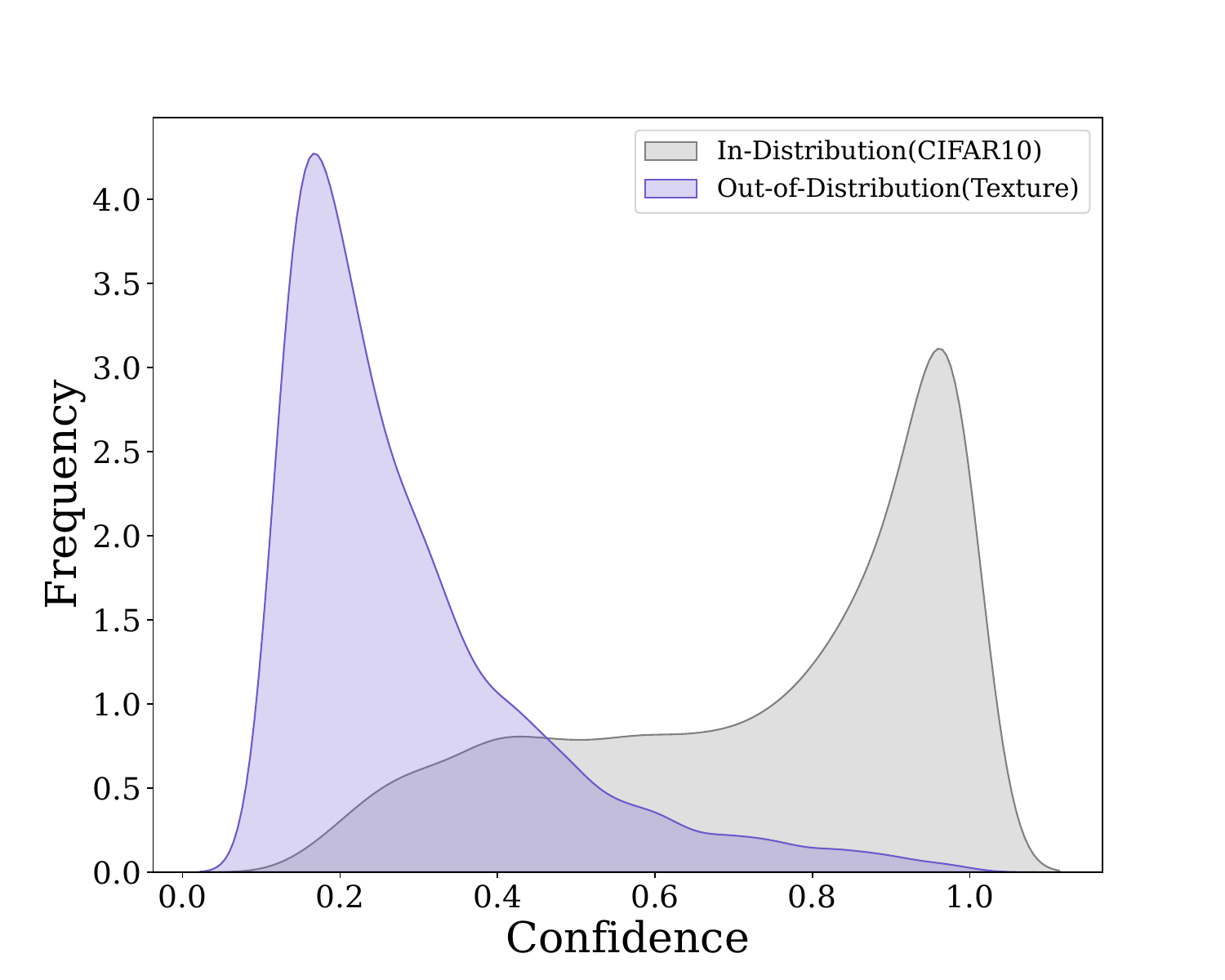}
            \includegraphics[scale=0.17]{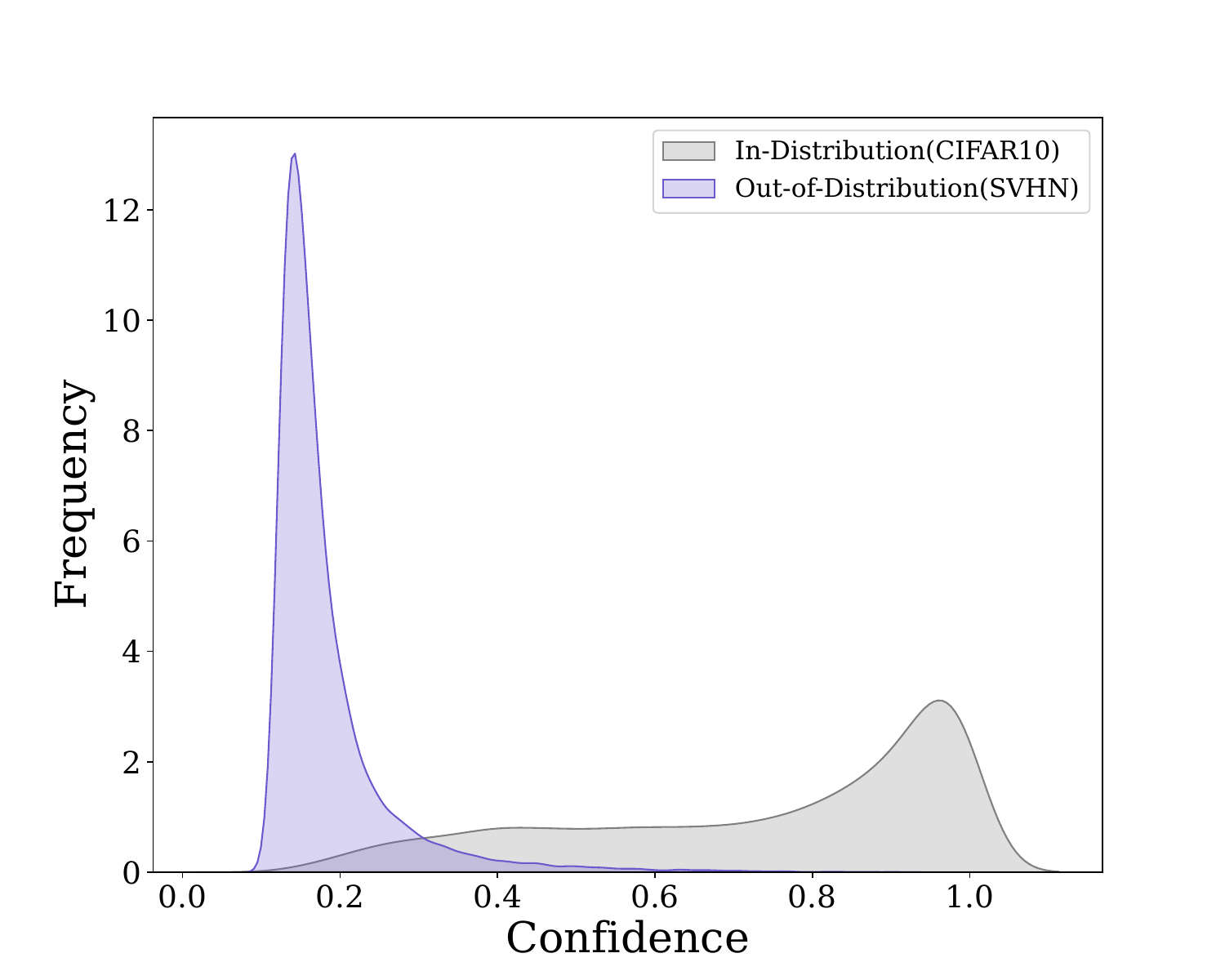}}\\
\subfloat{
		\includegraphics[scale=0.17]{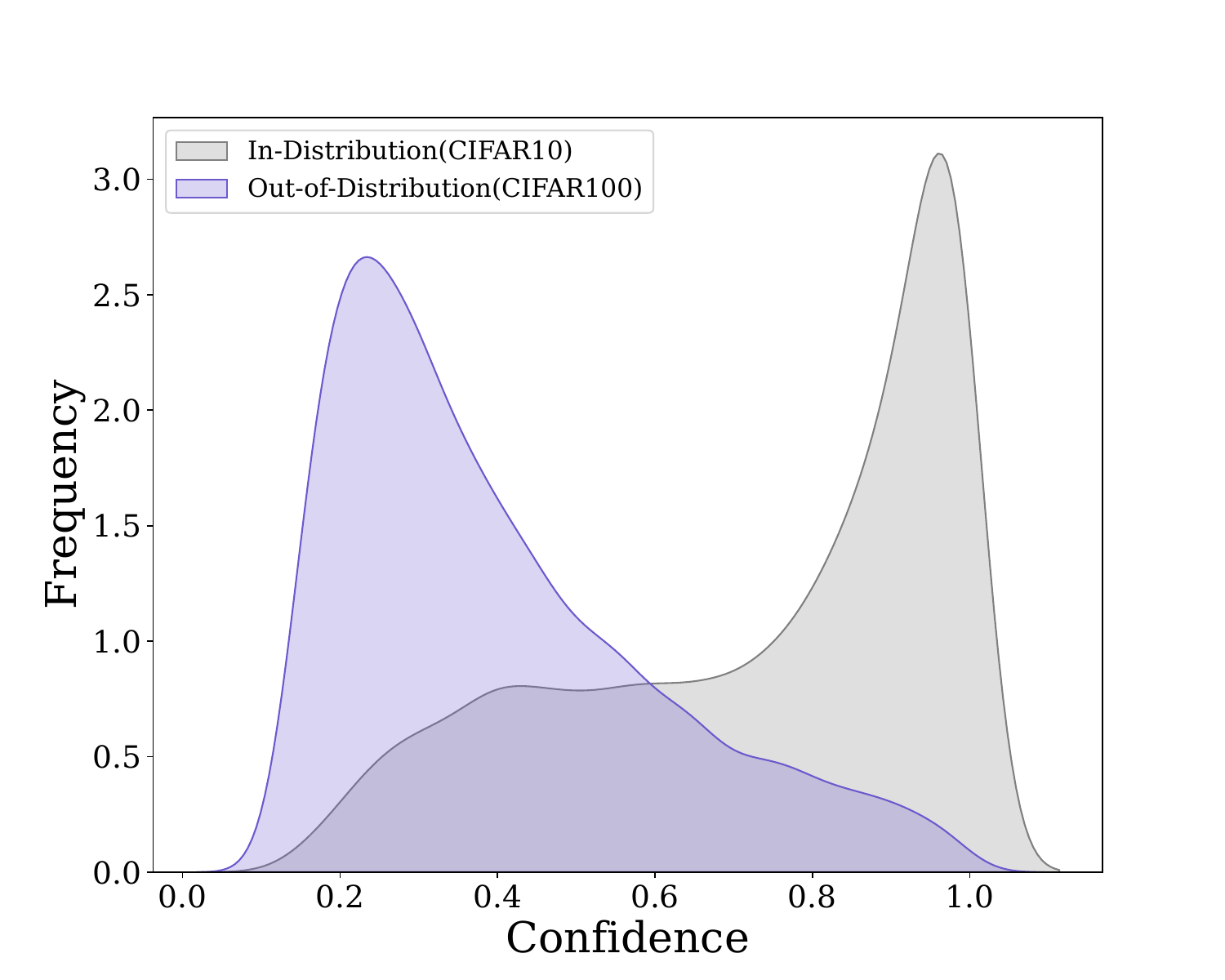}
            \includegraphics[scale=0.17]{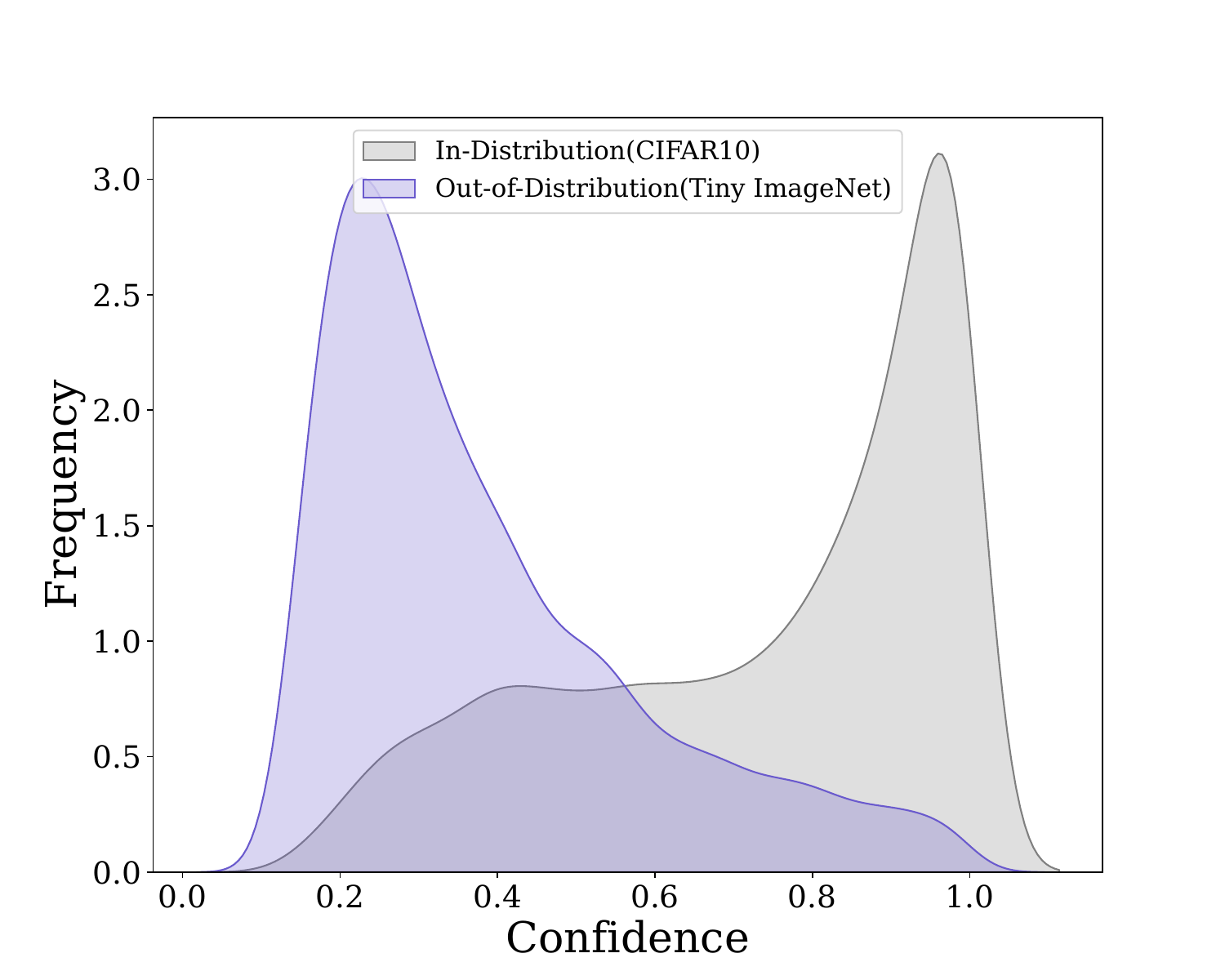}}\\
\subfloat{
		\includegraphics[scale=0.17]{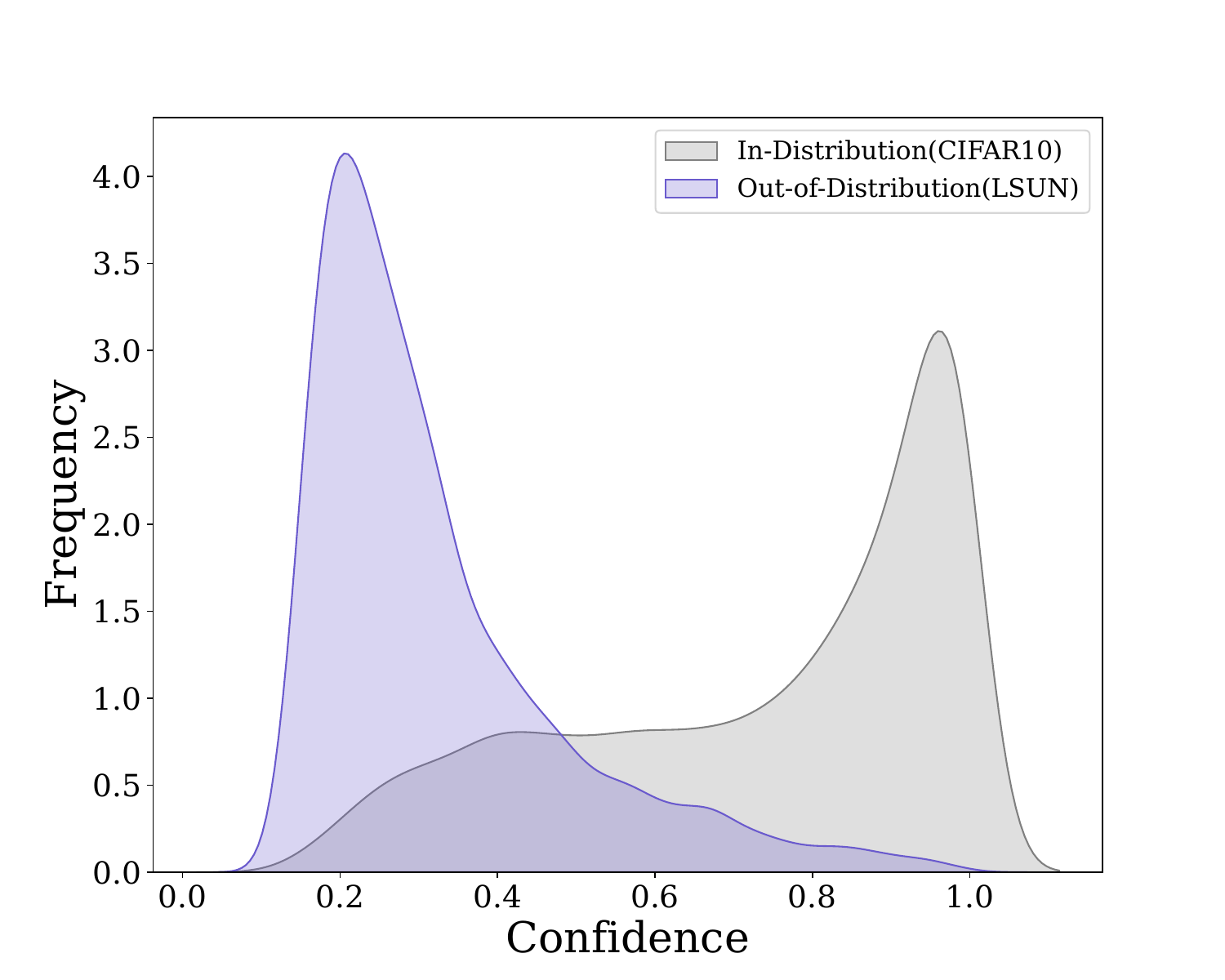}
            \includegraphics[scale=0.17]{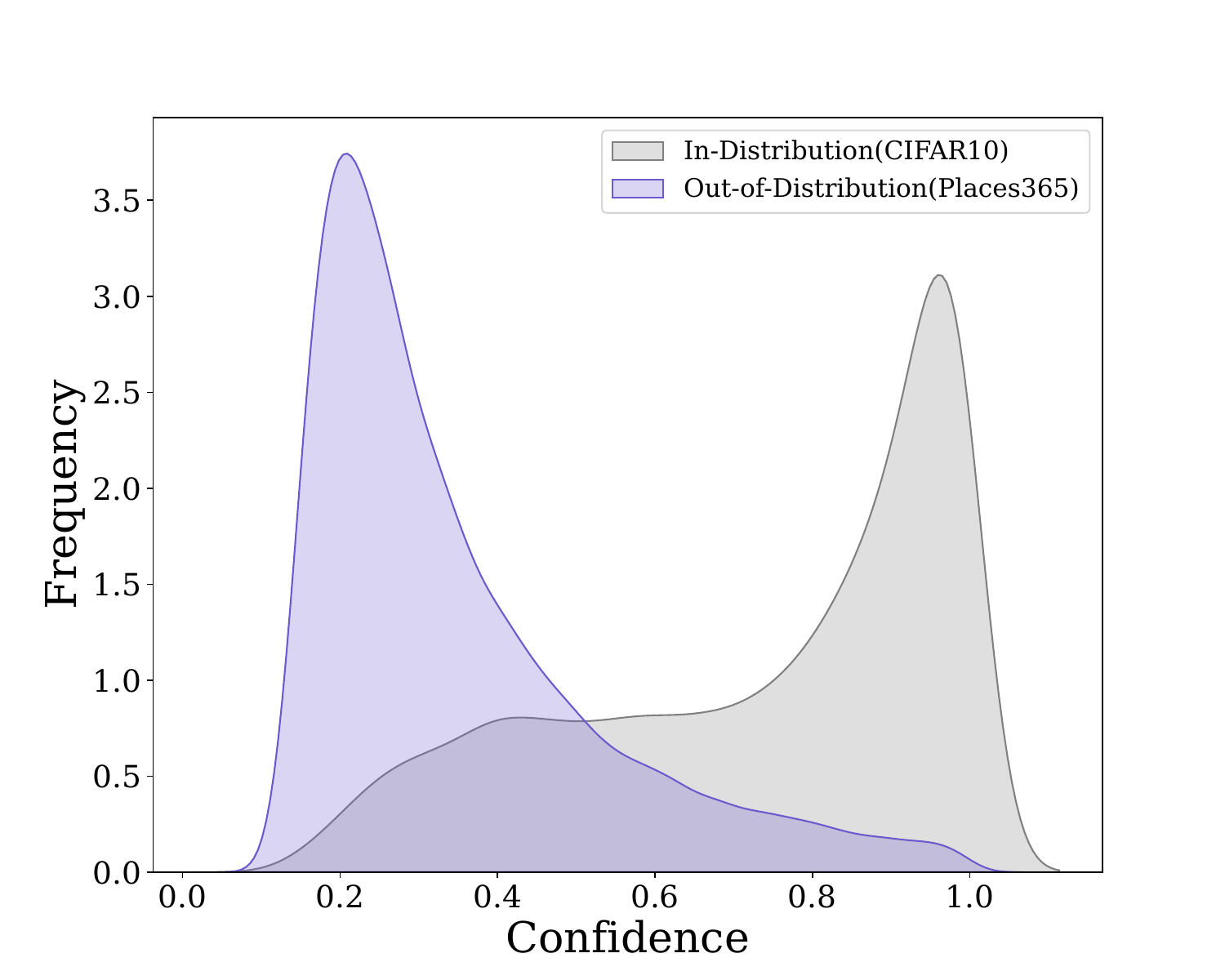}}
\caption{Confidence Distribution from CIFAR10-LT and other six test OOD datasets.}
\label{fig_density}
\end{figure}

\begin{figure}[!t]
    \centering
    \includegraphics[width=0.9\linewidth]{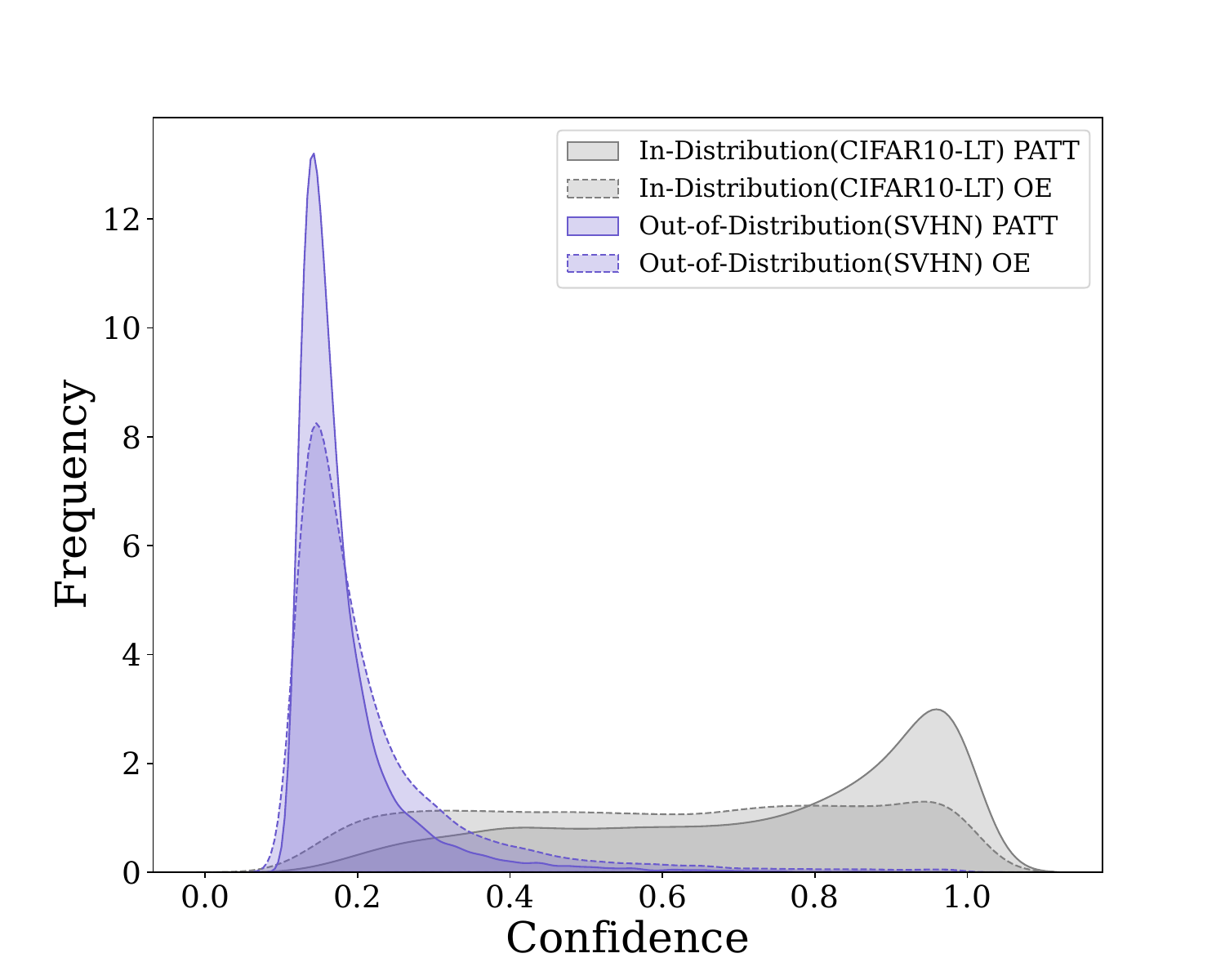}
    \caption{Visualization of confidence frequency from our method and OE. The CIFAR10 is used as ID data and SVHN is OOD test dataset.}
    \label{fig_density_compare}
\end{figure}

\begin{figure}[!t]
    \centering
    \includegraphics[width=0.9\linewidth]{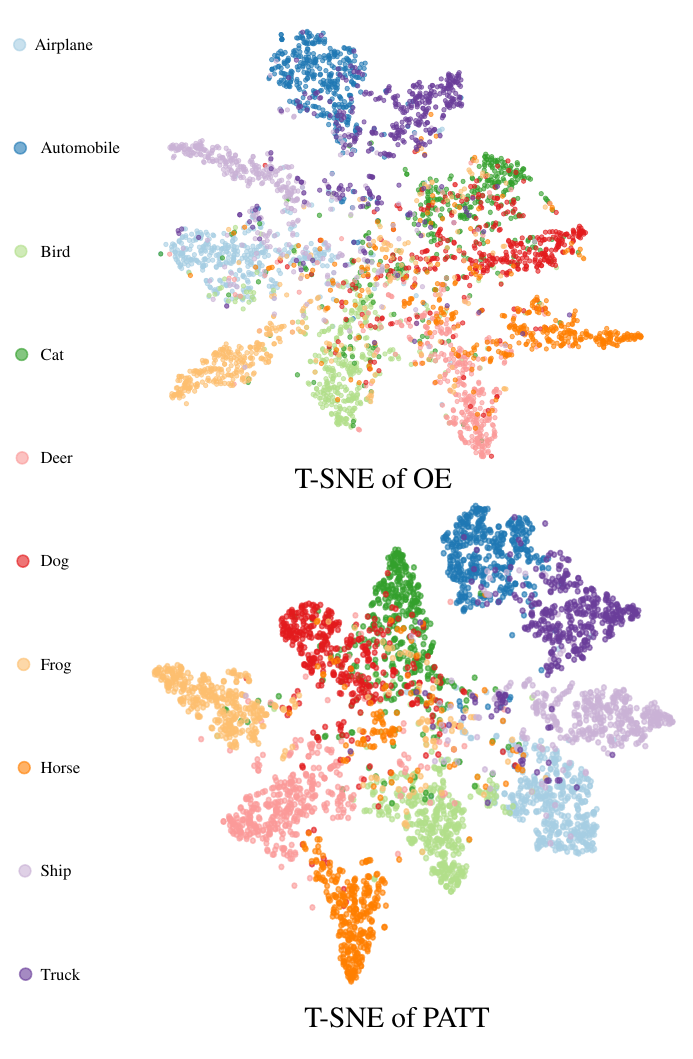}
    \caption{Visualization of in-distribution samples from CIFAR10 test set in the feature space. We visualize the penultimate layer features of ResNet18 trained by OE and PATT. 400 test samples from each class are visualized.}
    \label{tsne}
\end{figure}

\subsubsection{Interval Boundary}
We scale the attention weights to a range of [0,2] instead of [0,1] based on a simple idea that certain features should be amplified (falling within [1,2]) while others should be reduced (falling within [0,1]). Empirically, we conduct an experiment on CIFAR10-LT to compare the effectiveness of [0,2] and [0,1]. The results in the table below show a superiority of [0,2] over [0,1] across all OOD detection metrics, except that ID classification (ACC) is identical to [0,1]. These results verify that using [0,2] can yield better OOD detection performance.

\subsection{D.2 More Visualization}
\subsubsection{Confidence Frequency}
Figure~\ref{fig_density} illustrates the confidence distribution from CIFAR10-LT and six other test OOD datasets for CIFAR10-LT. It shows that both ID and OOD data naturally form smooth distributions, directly demonstrating the model's OOD detection capabilities. We can observe that on test OOD datasets with better OOD detection performance, the confidence for OOD data is closer to zero, resulting in less overlap with ID data. This clearly highlights the model's effectiveness in OOD detection. Additionally, the distinct separation of confidence levels between ID and OOD data ensures that the model can reliably identify outliers, even in complex and varied datasets.

To assess the confidence distribution of our method, Figure~\ref{fig_density_compare} illustrates the comparison of confidence distribution from CIFAR10-LT and SVHN between our method and baseline OE. We can see that our method significantly surpasses OE in ID confidence while being notably lower in OOD confidence. This indicates a much lower overlap probability between the two in our method compared to OE, which means better OOD detection performance.

\subsubsection{Feature Distribution}
Figure ~\ref{tsne} illustrates the penultimate layer features of ResNet18 from OE (above) and PATT (below). We can see that the feature distribution obtained by the OE model is difficult to distinguish between classes, and the feature space of the tail classes is significantly smaller than that of the head classes, such as "Ship" (slate-blue) and "Automobile" (blue). In contrast, the feature distribution obtained by PATT occupies a relatively balanced feature space, with a clear separation between classes. Only similar classes, such as Cat (green) and Dog (red), have some overlap, which is not caused by the long-tail distribution.

\begin{table}[!h]
\centering

\setlength{\tabcolsep}{1mm}
\begin{tabular}{ cc }
\toprule
 {Method} & {Time (s)} \\
\midrule
\midrule
 PASCL & 2140  \\
 EAT & 4248 \\
 PATT & \textbf{2078}\\
\midrule
\bottomrule
\end{tabular}
\caption{Average Computation Trade-off over 5 replay on CIFAR10-LT with different method.}
\label{tab:trade}
\end{table}

\subsubsection{Computation Trade-off}
Table~\ref{tab:trade} shows the total training time for each method on CIFAR10-LT using ResNet18 on an RTX-3090 GPU. Notably, PASCL and EAT are two-stage methods, whereas our PATT not only avoids additional training costs but also achieves end-to-end training.

\end{document}